\documentclass{article} 
\usepackage{iclr2027_conference,times}

\newcommand{\iclrarxiv}{%
  \iclrfinalcopy
  \AddToHook{cmd/maketitle/after}{%
    \lhead{Preprint}%
  }%
}

\usepackage[utf8]{inputenc} 
\usepackage[T1]{fontenc}    
\usepackage{hyperref}       
\usepackage{url}            
\usepackage{fontawesome5}   
\usepackage{booktabs}       
\usepackage{amsfonts}       
\usepackage{nicefrac}       
\usepackage{microtype}      
\usepackage{xcolor}         
\usepackage{amsmath}
\usepackage{graphicx}
\usepackage{multirow}
\usepackage{wrapfig}
\usepackage[skins,breakable]{tcolorbox}
\usepackage{textcomp}
\usepackage{xcolor}

\tcbuselibrary{listings}
\usepackage{xcolor}
\usepackage{textcomp}
\usepackage[skins,breakable]{tcolorbox}
\tcbuselibrary{listings}

\definecolor{gsmframeheader}{RGB}{128,128,128}
\definecolor{gsmframebackground}{RGB}{245,245,245}
\definecolor{gsmerrorbackground}{RGB}{255,225,225}
\definecolor{gsmerrortext}{RGB}{155,30,30}
\definecolor{gsmcorrectbackground}{RGB}{222,242,225}
\definecolor{gsmcorrecttext}{RGB}{28,100,49}

\newcommand{\gsmerror}[1]{%
  \begingroup
  \setlength{\fboxsep}{1pt}%
  \colorbox{gsmerrorbackground}{\textcolor{gsmerrortext}{\strut#1}}%
  \endgroup
}
\newcommand{\gsmcorrect}[1]{%
  \begingroup
  \setlength{\fboxsep}{1pt}%
  \colorbox{gsmcorrectbackground}{\textcolor{gsmcorrecttext}{\strut#1}}%
  \endgroup
}

\newtcolorbox{gsm8kframe}[3]{
  enhanced,
  breakable,
  colback=gsmframebackground,
  colframe=gsmframeheader,
  colbacktitle=gsmframeheader,
  coltitle=white,
  boxrule=0.5pt,
  arc=1.5mm,
  boxsep=0pt,
  left=5pt, right=5pt, top=3pt, bottom=5pt,
  toptitle=3pt, bottomtitle=3pt,
  titlerule=0pt,
  fontupper=\rmfamily\small,
  fonttitle=\rmfamily\small\bfseries,
  title={Sampling Steps: #1\hfill{\mdseries Accuracy: #2}},
  before upper={%
    \setlength{\parindent}{0pt}%
    \setlength{\parskip}{0pt}%
    \textbf{Question:} #3\par\smallskip
    \textbf{Generated answer:}\par\nobreak
  },
  before skip=8pt,
  after skip=8pt
}

\lstdefinestyle{gsmcode}{
  language=Python,
  basicstyle=\ttfamily\footnotesize,
  keywordstyle=\color{black},
  commentstyle=\color{black},
  stringstyle=\color{black},
  identifierstyle=\color{black},
  columns=fullflexible,
  keepspaces=true,
  tabsize=4,
  showspaces=false,
  showstringspaces=false,
  showtabs=false,
  showlines=true,
  upquote=true,
  numbers=none,
  breaklines=true,
  breakatwhitespace=false,
  breakautoindent=true,
  breakindent=1em,
  postbreak={},
  aboveskip=0pt,
  belowskip=0pt,
  mathescape=false,
  texcl=false,
  escapeinside={(*@}{@*)},
  escapebegin=\ttfamily\footnotesize,
  literate={’}{{\textquoteright}}1
           {‘}{{\textquoteleft}}1
           {“}{{\textquotedblleft}}1
           {”}{{\textquotedblright}}1
           {–}{{\textendash}}1
           {—}{{\textemdash}}1
           {…}{{\textellipsis}}1
           {−}{{\textminus}}1
           {×}{{\texttimes}}1
           {÷}{{\textdiv}}1
           {€}{{\texteuro}}1
           {£}{{\textsterling}}1
           { }{{~}}1
}

\newtcblisting{gsmcode}[4][]{
  enhanced,
  breakable,
  listing only,
  listing options={style=gsmcode},
  colback=gsmframebackground,
  colframe=gsmframeheader,
  colbacktitle=gsmframeheader,
  coltitle=white,
  boxrule=0.5pt,
  arc=1.5mm,
  boxsep=0pt,
  left=5pt, right=5pt, top=3pt, bottom=5pt,
  toptitle=3pt, bottomtitle=3pt,
  titlerule=0pt,
  fonttitle=\rmfamily\small\bfseries,
  title={Sampling Steps: #2\hfill{\mdseries Accuracy: #3\quad Executed answer: #4}},
  before skip=8pt,
  after skip=8pt,
  #1
}

\usepackage{algorithm}
\usepackage{listings}
\usepackage{caption}
\usepackage{mathtools}
\newcommand{\graycell}[1]{\textcolor{black!50}{#1}}

\definecolor{analysisblue}{RGB}{29,117,170}
\definecolor{analysisorange}{RGB}{250,125,14}
\definecolor{analysispurple}{RGB}{120,100,230}

\newcommand{\alg}[1]{Algorithm~\ref{alg:#1}}

\definecolor{codegreen}{RGB}{42,125,128}
\definecolor{codeblue}{RGB}{0,45,180}
\definecolor{codepink}{RGB}{220,40,130}
\definecolor{codeorange}{RGB}{220,95,0}

\lstdefinestyle{pseudocode}{
    language=Python,
    basicstyle=\ttfamily\small,
    keywordstyle=\color{codeblue}\bfseries,
    commentstyle=\color{codegreen}\bfseries,
    stringstyle=\color{codeorange},
    showstringspaces=false,
    columns=fullflexible,
    keepspaces=true
}

\title{Representation-Space MMD \\ for Diffusion Language Models}
\author{%
\begin{minipage}[t]{\dimexpr\textwidth-2\tabcolsep\relax}
\normalfont
\makebox[\linewidth][s]{%
  \textbf{Ilya Drobyshevskiy}$^{1,2,*}$\hfill
  \textbf{Ilia Sudakov}$^{1,2,*}$\hfill
  Maksim Semenov$^{2}$\hfill
  Denis Kuznedelev$^{1}$}\\
\begin{tabular*}{\linewidth}[t]{@{\extracolsep{\fill}}llr@{}}
  Maksim Ignatov$^{1}$ & Pavel Temirchev$^{1}$ & Nikita Balagansky$^{5}$\\
  Viacheslav Meshchaninov$^{6}$ & Nikita Gushchin$^{3,4}$ & \textbf{Dmitry Baranchuk}$^{1}$
\end{tabular*}\\[6pt]
\centering
\small
\makebox[\linewidth][s]{%
  $^1$Yandex Research\hfill
  $^2$HSE University\hfill
  $^3$Applied AI Institute\hfill
  $^4$AXXX\hfill
  $^5$T-Tech\hfill
  $^6$Constructor University}
\end{minipage}%
}
\iclrarxiv

\begin{document}
\maketitle
\begingroup
\renewcommand{\thefootnote}{\fnsymbol{footnote}}
\footnotetext[1]{Equal contribution.}
\endgroup

\begin{abstract}
We introduce a post-training method for diffusion language models (DLMs) that minimizes Maximum Mean Discrepancy (MMD) between generated and reference distributions in the feature space of a frozen pretrained DLM.
To estimate MMD, we retain contextual features at individual token positions, obtaining multiple observations per sequence from a single extractor pass.
We optimize this objective using policy gradients for discrete models and direct differentiation through generated latents for continuous models. 
In both cases, computing the loss directly from these features enables efficient post-training without full sampling trajectories or jointly trained auxiliary models.
Experiments show lower generative perplexity at comparable entropy on OpenWebText and better accuracy–computation trade-offs on GSM8K. 
On 16B DMax-LLaDA2.0 models with hybrid masked–uniform diffusion, we increase decoding parallelism with similar or higher accuracy on math and code benchmarks.
Code is available at \href{https://github.com/yandex-research/dlm-mmd}{\faGithub~\texttt{yandex-research/dlm-mmd}}.
\end{abstract}

\section{Introduction}

Diffusion language models (DLMs) generate text by refining noisy inputs in a discrete token space or a continuous latent space.
In discrete models, token-wise cross-entropy trains a factorized denoiser to fit conditional token marginals.
In continuous models, squared-error prediction of clean latents estimates their conditional mean.
These conditional estimates guide iterative sampling, but do not necessarily yield high-quality samples in a few steps.

Diffusion distillation addresses this challenge in several ways.
Some methods train a student to reproduce several teacher denoising updates in fewer steps~\citep{salimans2022progressive, luhman2021knowledge, deschenaux2024sdtt, kim2025cdlm}.
Others obtain distribution-matching gradients using an auxiliary model trained on the generator's evolving outputs~\citep{yin2024dmd,hoogeboom2026dmmd, zheng2026didi}.
These methods improve sampling efficiency but require teacher-derived targets or jointly trained auxiliary models.
Therefore, we ask whether a useful distributional training signal can instead be computed directly from generated and reference samples.

Maximum Mean Discrepancy (MMD) offers this possibility through kernel similarities between samples~\citep{gretton2012kernel}.
However, its effectiveness depends on the comparison space, since simple distances in high-dimensional spaces need not capture meaningful differences between outputs~\citep{li2017mmdgan, li2015gmmn}.
In this context, learned representations offer a promising comparison space for both discrete sequences and continuous outputs.
In visual generation, MMD and related objectives in pretrained feature spaces have enabled high-quality generation with one or a few steps~\citep{yang2026frechet, starodubcev2026swd,deng2026drifting}.
More recently, feature matching has improved LLM fine-tuning by comparing statistics of generated and reference sequences~\citep{jelassi2026matching}.
Together, these results suggest that a fixed pretrained feature space can provide an effective training signal for distribution matching.



\paragraph{Contributions.} 
In this work, we explore MMD for post-training both discrete and continuous DLMs.
We use a frozen pretrained DLM to extract contextual features at individual positions, obtaining multiple observations from each generated or reference sequence in a single extractor pass.
We then train the generator to match the distributions of these features.

For discrete models, we sample sequences from the token distributions produced by a single denoiser pass and optimize the resulting MMD reward using REINFORCE with a leave-one-out baseline.
We instantiate it as MDLM-MMD for masked-token prediction~\citep{sahoo2024simple} and as DMax-MMD for hybrid masked--uniform diffusion~\citep{chen2026dmaxaggressiveparalleldecoding}, where MMD trains both masked prediction and refinement of the model's own token predictions.

For continuous DLMs (CDLMs), we initialize a generator from a pretrained Embedded Language Flows (ELF) model~\citep{hu2026elf} and train it to map Gaussian noise to clean latent sequences in a single step.
In this case, MMD gradients directly propagate through the frozen ELF feature extractor and the generated latents.
At inference, we use self-conditioning to iteratively refine the single-step predictions, extending the generator to multi-step sampling.

We evaluate both realizations on OpenWebText and TinyGSM, where they outperform the evaluated distribution-matching and flow-map baselines in most setups.
At the 16B scale, we show that DMax-MMD improves the accuracy-efficiency trade-off on established math and code benchmarks.


\section{Preliminaries}
\label{sec:background}
\vspace{-1mm}

In this section, we briefly review discrete and continuous DLMs and introduce MMD, which forms the basis of our post-training objective.
A broader discussion of related work is provided in App.~\ref{app:related_work}.
Throughout, $\mathbf{x}$ denotes a token sequence for discrete DLMs
or a sequence of latent vectors for continuous DLMs.
We use $t=0$ for maximal corruption and $t=1$ for clean data.

\textbf{Discrete DLMs} corrupt a clean sequence by independently retaining each token with probability $\alpha_t$ and otherwise drawing a replacement from a fixed noise distribution~\citep{austin2021d3pm}.
Given the corrupted state $\mathbf{x}_t$, the denoiser
$q_\theta(\mathbf{x}\mid\mathbf{x}_t)$ predicts a factorized distribution over clean tokens, so tokens can be sampled independently from a single forward pass.

In masked diffusion, the noise distribution is concentrated on an absorbing \texttt{[MASK]} symbol, and the denoiser preserves visible tokens~\citep{sahoo2024simple,shi2024simplified}.
Pretraining minimizes weighted cross-entropy over masked positions.
At inference, generation starts from a fully masked sequence and progressively reveals tokens~\citep{nie2025llada}.
We use confidence-threshold decoding~\citep{fastdllm},
which adaptively determines how many tokens are revealed per step.

Uniform diffusion instead draws replacement tokens uniformly from the vocabulary and applies denoising supervision to all positions.
Sampling typically starts from a uniformly random sequence and repeatedly updates all positions, allowing earlier predictions to be revised~\citep{austin2021d3pm}.

\paragraph{Continuous DLMs.}
In this work, we build on the recent Embedded Language Flows (ELF)~\citep{hu2026elf},
which generate text in the latent space of a frozen encoder.
The encoder maps a token sequence $\mathbf{s}$ to latents
$\mathbf{x}=E(\mathbf{s})\in\mathbb{R}^{L\times d}$.
Then, ELF learns to predict clean latents from interpolated states
$\mathbf{z}_t=t\mathbf{x}+(1-t)\boldsymbol{\epsilon}$,
where $\boldsymbol{\epsilon}\sim\mathcal{N}(\mathbf{0},\mathbf{I})$,
using flow matching~\citep{lipman2023flow}.

To inform each prediction, ELF uses \textit{self-conditioning}
($\mathrm{SC}$), which supplies the denoiser with a previous
estimate of the clean latents~\citep{chen2023analogbits}.
When no estimate is available, this input is $\mathbf{0}$.
During standard sampling, the model's predictions define a
velocity field that is integrated from noise toward clean latents,
with each prediction providing self-conditioning for the next step.
The same network then decodes the final latent sequence into tokens.

\paragraph{Maximum Mean Discrepancy.}
For distributions $P$ and $Q$ and a positive-definite kernel $k$, squared MMD is~\citep{gretton2012kernel}
\begin{equation}
\label{eq:mmd-loss}
    \operatorname{MMD}^2_k(P,Q)
    =
    \textcolor{analysisblue}{
        \mathbb{E}_{\mathbf{x},\mathbf{x}'\sim P}\!\left[k(\mathbf{x},\mathbf{x}')\right]
    }
    +\textcolor{analysisorange}{
        \mathbb{E}_{\mathbf{y},\mathbf{y}'\sim Q}\!\left[k(\mathbf{y},\mathbf{y}')\right]
    }
    -\textcolor{analysispurple}{
        2\mathbb{E}_{\mathbf{x}\sim P,\mathbf{y}\sim Q}\!\left[k(\mathbf{x},\mathbf{y})\right]
    },
\end{equation}
where the draws in each expectation are independent.
The linear kernel $k(\mathbf{a},\mathbf{b})=\mathbf{a}^{\top}\mathbf{b}$ matches feature means, giving $\operatorname{MMD}^2_k(P,Q)=\|\boldsymbol{\mu}_P-\boldsymbol{\mu}_Q\|_2^2$.
The Gaussian RBF kernel $k(\mathbf{a},\mathbf{b})=\exp(-\|\mathbf{a}-\mathbf{b}\|_2^2/(2\sigma^2))$, with $\sigma>0$, is characteristic, so zero MMD identifies equal distributions in the space where the kernel is evaluated.
When $Q$ is the generated distribution, its within-distribution term penalizes average RBF similarity between generated features, while the cross term rewards similarity to reference features~\citep{li2015gmmn,arbel2019maximum}.

\section{Method}
\label{sec:method}

\begin{figure*}[t!]
\centering
\includegraphics[width=\linewidth]{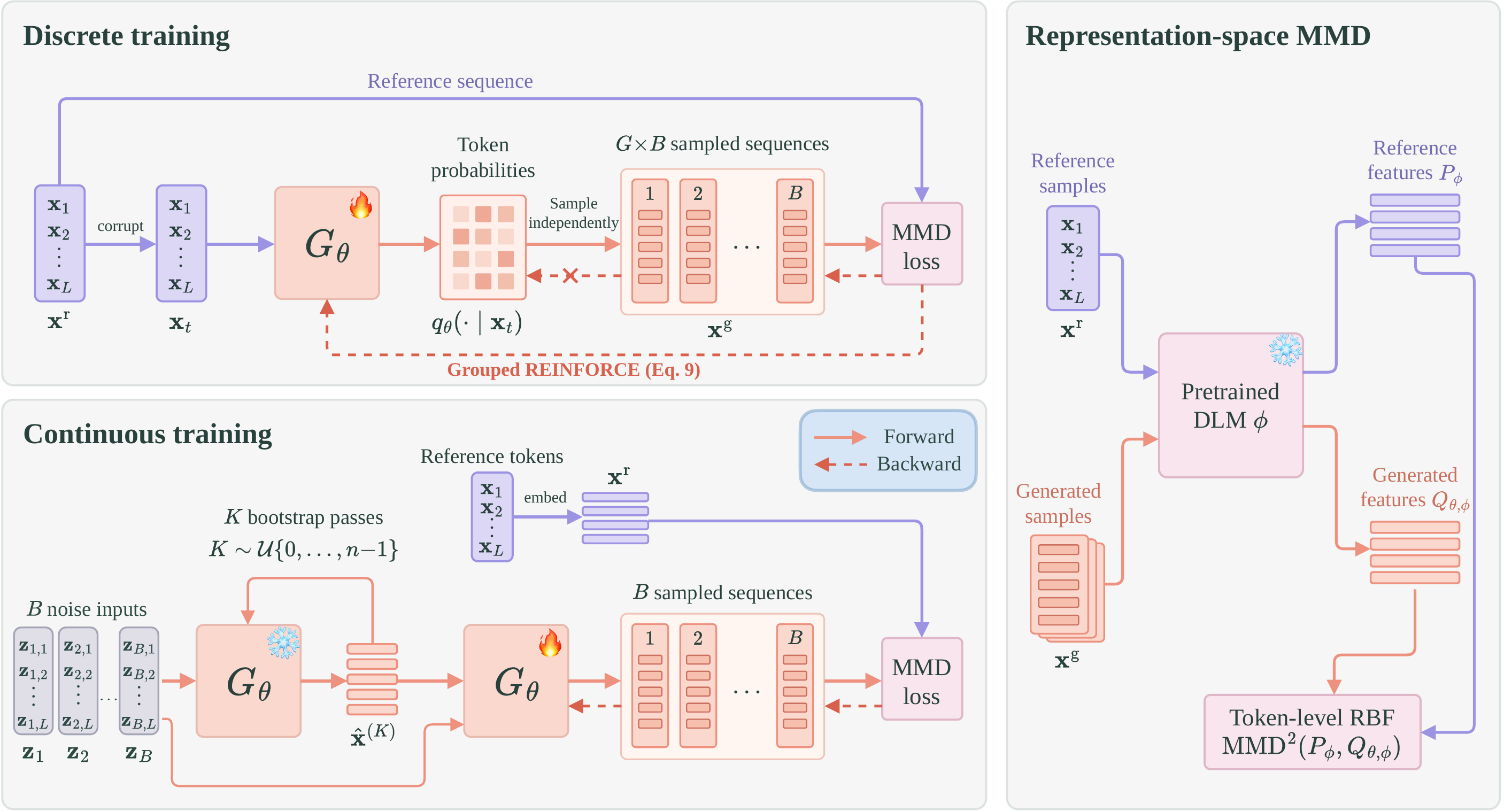}
\caption{\textbf{Representation-space MMD training.}
We illustrate the case of one reference sequence and $B$ generated sequences per MMD estimate.
For discrete DLMs (\textit{top left}), we independently draw $G$ such batches from the token distributions produced by a single denoiser pass and optimize the loss with grouped REINFORCE.
For continuous DLMs (\textit{bottom left}), $K$ bootstrap passes provide self-conditioning without gradients, followed by a differentiable generation pass.
In both cases, token-level RBF MMD compares contextual features extracted by a frozen pretrained DLM (\textit{right}).}

\label{fig:main_scheme}
\end{figure*}


We post-train a DLM by matching contextual token representations of generated and reference sequences.
Using the features at individual positions gives us multiple samples for MMD from each extractor evaluation.
We first describe the objective and its estimation, then turn to the discrete and continuous realizations.

\subsection{Representation-space MMD Training}
\label{sec:training}

\paragraph{Choosing a representation space.}
To compare generated and reference samples with MMD, we need a representation in which to measure their similarity.
A pretrained DLM already computes features that describe tokens in their available context, making it a natural choice of feature extractor.
Therefore, we use a frozen pretrained DLM and collect its intermediate hidden states,
\begin{equation}
\phi(\mathbf{x})
=\bigl(\phi_1(\mathbf{x}),\ldots,\phi_L(\mathbf{x})\bigr)
\in\mathbb R^{L\times D},
\end{equation}
where $L$ is the sequence length and $D$ is the feature dimension.

These contextual features let comparisons at individual positions reflect the surrounding sequence without requiring an externally trained feature model.

In practice, we extract features from clean samples, as corruption at intermediate diffusion timesteps did not improve performance in our experiments.
We apply the same extraction rule to real and generated samples and explore alternative extractors in App.~\ref{app:encoders}.

\paragraph{Using token features for distribution matching.}
Pooling the features into a single embedding would leave only one observation per sequence.
Instead, we retain the individual representations, obtaining $|S|$ feature samples in one forward pass, where $S$ is the nonempty set of scored positions.
This provides more observations for MMD estimation, which is especially useful when only a single reference is available for a condition.

We score all positions for unconditional latent generation, response positions for prompted latent generation, and masked positions for masked DLMs.
The extractor still processes the available context, so the selected features can carry information from outside $S$.
Features within a sequence can therefore be dependent, which we account for in the estimator below.

To define their distribution, draw $\mathbf{x}\sim p_{\mathrm{data}}(\cdot\mid c)$ and choose $J$ uniformly from $S$.
Here, $c$ is the generation condition, which may be empty, a prompt, or a corrupted state.
We denote the distribution of $\phi_J(\mathbf{x})$ by $P_\phi(\cdot\mid c)$ and define $Q_{\theta,\phi}(\cdot\mid c)$ analogously using generated sequences.
We then match these distributions with the Gaussian RBF kernel $k$ from Section~\ref{sec:background},
\begin{equation}
\mathcal L_{\mathrm{MMD}}(\theta)
=\mathbb E_c\left[
\operatorname{MMD}_k^2\!\left(
P_\phi(\cdot\mid c),Q_{\theta,\phi}(\cdot\mid c)
\right)\right].
\label{eq:conditional_mmd}
\end{equation}
Beyond providing a training signal, feature-space MMD can in principle identify the underlying data distribution.
As used in MMD-GAN~\citep{li2017mmdgan}, applying a characteristic kernel through an injective feature map gives zero population MMD if and only if the original distributions agree.
Recent results on causal transformers show that even a last-token representation can be almost surely injective in the discrete input sequence under the stated assumptions~\citep{nikolaou2026languagemodelsinjectiveinvertible}.
These results motivate our choice of DLM token features, but do not establish that matching our token-feature distributions uniquely identifies the sequence distribution.

\paragraph{Estimating the training objective.}
Although the distribution above is defined by sampling one position, a forward pass provides features at every scored position.
We use all of them by averaging kernel similarities between two sequences,
\begin{equation}
\kappa_\phi(\mathbf{x},\mathbf{x}')
=\frac{1}{|S||S'|}\sum_{i\in S}\sum_{j\in S'}
k\!\left(\phi_i(\mathbf{x}),\phi_j(\mathbf{x}')\right),
\label{eq:token_average_kernel}
\end{equation}
where $S$ and $S'$ are their scored-position sets.
This average is the inner product of empirical kernel mean embeddings and hence a positive-semidefinite kernel on sequences~\cite[Eq.~(3.40-3.41)]{muandet2020kernelmeanembeddingdistributions}.

We first describe MMD estimation for a single batch in the unconditional setting. 
The training objective averages these estimates over batches.
MMD requires independent draws in each kernel expectation.
Since features within a sequence can be dependent, we form the within-distribution comparisons using different sequences.
Consider real and generated batches
$\mathbf X^{\mathrm r}=(\mathbf x_b^{\mathrm r})_{b=1}^B$ and
$\mathbf X^{\mathrm g}=(\mathbf x_b^{\mathrm g})_{b=1}^B$,
with i.i.d.\ sequences in each batch and independence between batches.
For a fixed extractor and kernel and $B\geq2$, the usual MMD construction~\citep{gretton2012kernel} gives
\begin{equation}
\begin{aligned}
\widehat{\mathcal L}_{\mathrm{MMD}}(\theta)
={}&\frac{1}{B(B-1)}\sum_{b\ne b'}
\left[
\textcolor{analysisblue}{\kappa_\phi(\mathbf x_b^{\mathrm r},\mathbf x_{b'}^{\mathrm r})}
+\textcolor{analysisorange}{\kappa_\phi(\mathbf x_b^{\mathrm g},\mathbf x_{b'}^{\mathrm g})}
\right]-\frac{2}{B^2}\sum_{b,b'}
\textcolor{analysispurple}{\kappa_\phi(\mathbf x_b^{\mathrm r},\mathbf x_{b'}^{\mathrm g})}.
\end{aligned}
\label{eq:token_mmd_estimator}
\end{equation}
The restriction $b\ne b'$ excludes all same-sequence token pairs, including pairs at different positions, to preserve unbiasedness.
Thus, $B$ sequences with $|S|$ scored positions each contribute $B|S|$ feature observations without being treated as $B|S|$ independent draws.
Retaining same-sequence pairs generally introduces bias, although these comparisons may still be useful in practice.

The \textcolor{analysispurple}{real--generated term} encourages similarity to reference features, while the \textcolor{analysisorange}{generated--generated term} penalizes similarity among generated features.
Together, they balance attraction to the reference distribution against concentration around a few representations.
The \textcolor{analysisblue}{real--real term} is independent of $\theta$ and is omitted during optimization.

The \textcolor{analysisorange}{generated--generated term} requires $B\geq2$.
With $B=1$, retaining only the \textcolor{analysispurple}{real--generated term} gives an attraction-only objective, which leads to worse performance on discrete models and rapid collapse in continuous models in our ablation study (Section~\ref{sec:exp_ablations}).

For conditional cases, which are more prevalent in practice, we can train with one reference per condition because the \textcolor{analysisblue}{real--real term} need not be estimated for optimization.
Given $\mathbf{x}^{\mathrm r}\sim p_{\mathrm{data}}(\cdot\mid c)$ and $B\geq2$ generated sequences sampled independently of one another, we estimate the remaining terms using the training loss
\begin{equation}
\widehat{\mathcal L}_{\mathrm{train}}(\theta\mid c)
=
\frac{1}{B(B-1)}
\sum_{b\ne b'}
\textcolor{analysisorange}{
\kappa_\phi(\mathbf{x}_b^{\mathrm g},\mathbf{x}_{b'}^{\mathrm g})}
-\frac{2}{B}\sum_b
\textcolor{analysispurple}{
\kappa_\phi(\mathbf{x}^{\mathrm r},\mathbf{x}_b^{\mathrm g})}.
\label{eq:conditional_training_loss}
\end{equation}
Its expectation over both reference and generated samples equals the conditional squared MMD up to an additive constant independent of $\theta$.
We average this loss over conditions, using one reference and $B$ generated sequences for each.

The following subsections describe feature extraction and optimization in each state space.
Fig.~\ref{fig:main_scheme} illustrates the training scheme for both discrete and continuous realizations.

\subsection{Discrete Diffusion Language Models}
\label{sec:discrete}

Given a noisy input $\mathbf{x}_t$, we select a nonempty set $S$ of valid positions to predict, restricted to the response for prompted tasks.
The denoiser defines a conditional distribution by predicting these positions while keeping the remaining input fixed,
\begin{equation}
q_\theta(\mathbf{x}^{\mathrm g}\mid\mathbf{x}_t)
=\prod_{i\in S}q_\theta(\mathbf{x}_i^{\mathrm g}\mid\mathbf{x}_t),
\qquad
\mathbf{x}_i^{\mathrm g}=\mathbf{x}_{t,i}\quad\text{for }i\notin S.
\label{eq:discrete_sample}
\end{equation}
We compare these sampled sequences with a reference $\mathbf{x}^{\mathrm r}$ that contains the clean target tokens at $S$ and shares the same context elsewhere.
Both are passed through the frozen feature extractor, and their features at $S$ enter the MMD loss.
Although the sampled tokens are conditionally independent, their contextual representations allow the loss to assess them jointly.

\paragraph{Optimizing sampled sequences.}
We use the conditional MMD estimate in Eq.~\eqref{eq:conditional_training_loss}, which compares one reference with $B\geq2$ independently sampled sequences.
We draw all $B$ sequences independently from the token distributions produced by a single denoiser forward pass.
Categorical sampling prevents direct differentiation through these samples, so we optimize the loss with REINFORCE~\citep{williams1992reinforce}.
To reduce gradient variance, we draw a group of $G$ independent MMD batches, each containing $B$ sampled sequences, for the same reference and corrupted state $\mathbf{x}_t$.
Each batch $\mathbf X_g=(\mathbf{x}_{g,1},\ldots,\mathbf{x}_{g,B})$ receives the negative MMD estimate as its reward.
For $G>1$, the other batches provide a leave-one-out baseline~\citep{ahmadian-etal-2024-back},
\begin{equation}
r_g=-\widehat{\mathcal L}_{\mathrm{train}}
       (\mathbf{x}^{\mathrm r},\mathbf X_g\mid\mathbf{x}_t),
\qquad
A_g=r_g-\frac{1}{G-1}\sum_{h\ne g}r_h.
\label{eq:discrete_reward}
\end{equation}
We then minimize the policy-gradient surrogate
\begin{equation}
\mathcal L_{\mathrm{PG}}
=-\frac{1}{G}\sum_{g=1}^{G}
A_g
\sum_{b=1}^{B}\sum_{i\in S}
\log q_\theta(\mathbf{x}_{g,b,i}\mid\mathbf{x}_t).
\label{eq:discrete_surrogate}
\end{equation}
Because a reward depends on an entire batch, it multiplies the sum of log probabilities of all sequences in that batch.
The baseline uses independent batches conditional on the shared reference and $\mathbf{x}_t$, so it does not change the expected gradient.
For $G=1$, we use $A_1=r_1$.

\paragraph{Masked and uniform variants.}
We consider two realizations of this training procedure.
In \textbf{MDLM-MMD}, $\mathbf{x}_t$ is formed by masking tokens and $S$ contains only masked positions, so visible tokens remain unchanged.
In \textbf{DMax-MMD}, we post-train DMax~\citep{chen2026dmaxaggressiveparalleldecoding}, a hybrid masked--uniform model, using MMD for both its masked and prediction losses.
The masked loss uses the same choice of $S$, while the prediction loss uses all target positions and conditions on the model's own token predictions.
For block diffusion, we apply the objective within each block conditioned on its available context.
After MMD post-training, we use each discrete model's original sampling procedure without any changes.

\subsection{Continuous Diffusion Language Models}
\label{sec:continuous}

Continuous latents allow us to optimize the MMD objective by differentiating through generated samples.
We instantiate this approach as \textbf{ELF-MMD}, initializing $G_\theta$ from a pretrained ELF model and training a \textit{one-step} latent generator,
\begin{equation}
\mathbf{x}^{\mathrm g}
=G_\theta\!\left(\mathbf z,t=0,\mathrm{SC}=\mathbf 0\right),
\qquad
\mathbf z\sim\mathcal N(\mathbf 0,\mathbf I).
\label{eq:one_step_generator}
\end{equation}
Here, $\mathrm{SC}$ denotes the self-conditioning input described in Section~\ref{sec:background}.

The pretrained ELF model provides the intermediate features used by MMD.
We feed each clean real or generated latent sequence $\mathbf x$ into both its latent and self-conditioning inputs at $t=1$,
\begin{equation}
\phi(\mathbf x)
=\phi_{\mathrm{ELF}}\!\left(
\mathbf x,t=1,\mathrm{SC}=\mathbf x
\right).
\label{eq:elf_features}
\end{equation}
We use Eq.~\eqref{eq:token_mmd_estimator} for unconditional generation and Eq.~\eqref{eq:conditional_training_loss} for prompted generation, where the extractor receives the clean prompt and only response positions enter the kernel.
The resulting gradients pass through both extractor inputs to update $G_\theta$.

\paragraph{Iterative refinement through self-conditioning.}
\label{sec:iterative_refinement}
To extend the obtained model to multi-step generation, we iteratively refine its predictions via self-conditioning.
Specifically, starting from $\hat{\mathbf x}^{(0)}=\mathbf 0$, we feed each prediction back into the generator,
\begin{equation}
\hat{\mathbf x}^{(k)}
=G_\theta\!\left(
\mathbf z^{(k)},t=0,\mathrm{SC}=\hat{\mathbf x}^{(k-1)}
\right),
\qquad k=1,\ldots,K.
\label{eq:iterative_refinement}
\end{equation}
The noise is either held fixed across iterations or independently resampled, while the diffusion time remains at $t=0$.
Thus, $K=1$ recovers the one-step generator.
After $K$ refinement steps, one final evaluation at $t=1$, with null self-conditioning, converts $\hat{\mathbf x}^{(K)}$ to token logits.
Taking their argmax yields the generated text, for a total of $K+1$ network evaluations (Algorithm~\ref{alg:ours_sampling}).

\paragraph{Bootstrapping and iterative refinement distillation.}
\label{sec:bootstrap}
Later refinement steps receive the model's previous predictions as self-conditioning, which differ from the null input used for one-step generation.
To expose the model to these inputs during training, we uniformly sample the number of preliminary refinement steps from $\{0,\ldots,n-1\}$, where $n$ is the maximum number of generator passes including the final differentiable step.
We run these preliminary steps with fixed noise and without gradients, then apply MMD to one additional prediction and backpropagate only through this final step.

In addition, to reduce the number of evaluations needed at inference, we optionally apply iterative refinement distillation (IRD) after MMD training.
A frozen ELF-MMD teacher supplies the final prediction of a multi-step trajectory, which a student learns to reproduce in one step~\citep{yoo2026fixedpoint,luhman2021knowledge}.
We describe the distillation objective in App.~\ref{sec:ird}.

We note that the MMD loss is also applicable to the continuous DLM variants that do not use self-conditioning~\citep{lee2026fmlm, deschenaux2026language}.
Instead of iterative refinement, they could be adapted to use alternative multi-step sampling schemes, such as consistency sampling~\citep{song2023consistency}, that are widely explored for visual generators~\citep{sauer2024add, yin2024dmd2}.

\section{Experiments}
\label{sec:experiments}
In this section, we evaluate whether representation-space MMD improves generation at limited sampling budgets for both discrete and continuous DLMs. 
We first study unconditional generation on OpenWebText and mathematical reasoning on GSM8K, then test the approach on 16B DMax models. 

\subsection{Unconditional Generation}
\label{sec:exp_uncond}

For unconditional generation, we follow prior work~\citep{hu2026elf,lee2026fmlm,chen2026langflow} and train on OpenWebText (OWT)~\citep{Gokaslan2019OpenWeb} with sequences packed to length $L=1024$.
We evaluate $1{,}000$ generated samples using generative perplexity (Gen. PPL) under pretrained GPT-2 Large~\citep{radford2019gpt2} and average unigram entropy.
We use the data entropy ($H \approx 5.43$) as a reference, rather than treating higher entropy as uniformly better.

\begin{figure}[!t]
\centering
\vspace{-2mm}
\includegraphics[width=\linewidth]{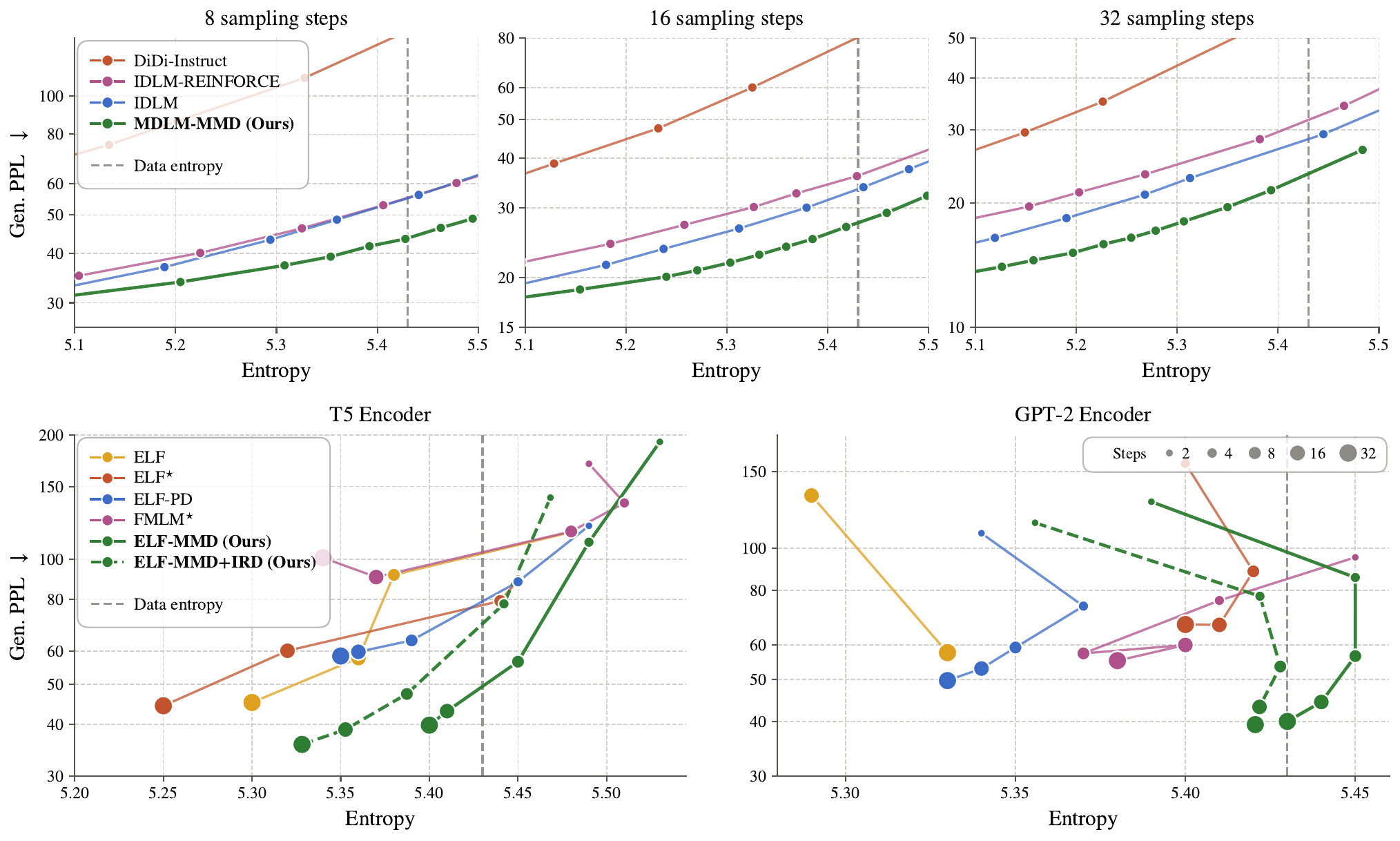}
\vspace{-5mm}
\caption{
    \textbf{Gen. PPL--Entropy trade-offs on OpenWebText.}
    \textbf{Top:} MDLM-based models with temperature sweeps at $8$, $16$, and $32$ sampling steps.
    \textbf{Bottom:} Continuous models evaluated at different sampling budgets.
    Low-entropy ELF and ELF$^{\star}$ points are omitted.
    \textit{Note:} entropy above the reference data entropy ($H {\approx}5.43$) does not imply better performance.
}
\label{fig:ppl_entropy_owt}
\vspace{-3mm}
\end{figure}

\paragraph{Discrete DLMs.}
We initialize MDLM-MMD from a pretrained MDLM~\citep{sahoo2024simple} and compute MMD rewards using features from a frozen MDLM feature extractor, with policy-gradient group size $G=4$.
We compare against DiDi-Instruct~\citep{zheng2026didi}, IDLM~\citep{li2026idlm}, and IDLM-REINFORCE, which optimizes the IDLM log-ratio reward using policy gradients. This comparison helps distinguish the effects of the training objective and the optimization procedure.
We evaluate each method at $8$, $16$, and $32$ sampling steps, varying the sampling temperature to characterize the trade-off between generative perplexity and sample entropy.
As shown in Fig.~\ref{fig:ppl_entropy_owt}, MDLM-MMD achieves lower generative perplexity than all baselines at matched entropy.
In particular, interpolating the curves at the reference-data entropy gives ${\sim}17$--$21\%$ lower generative perplexity than IDLM across the three sampling budgets. Exact numbers are reported in Tab.~\ref{tab:discrete_owt_resutls}.

\paragraph{Continuous DLMs.}
We use the ELF-B architecture with two choices of continuous latent space: the final-layer hidden representations of a T5-small encoder and a GPT-2 Large model, following~\citep{hu2026elf,yoo2026fixedpoint}.
For both OWT and GSM8K, we report ELF-MMD, obtained by MMD post-training alone, and ELF-MMD$+$IRD (App.~\ref{sec:ird}).
Fig.~\ref{fig:ppl_entropy_owt} compares both variants with ELF and its self-conditioning distilled variant ELF$^{\star}$~\citep{yoo2026fixedpoint}.
We also include progressively distilled ELF (ELF-PD)~\citep{hu2026elf} and FMLM$^{\star}$~\citep{yoo2026fixedpoint} as few-step baselines.


ELF-MMD improves generation across most sampling budgets for both T5 and GPT-2 encoders. At 8 steps, it achieves lower generative perplexity and entropy closer to the reference value than ELF-PD; at 32 steps, it outperforms ELF on both metrics. IRD further improves few-step results: at 4 steps, it reduces generative perplexity relative to ELF-MMD by \({\sim}30\) and \({\sim}10\) points with the T5 and GPT-2 encoders, respectively, while bringing entropy closer to the reference value for both encoders. Exact numbers, together with additional results for MMD post-training of ELF$^\star$, are reported in Tab.~\ref{tab:owt_resutls}.



\subsection{Conditional Generation}
\label{sec:exp_cond}

For conditional generation, following~\citep{kim2025train,agarwal2026posterior}, we train on TinyGSM~\citep{liu2023tinygsm} and evaluate final-answer accuracy on the GSM8K test set~\citep{cobbe2021training}. 
We pack sequences to a maximum length of $L=512$.
In addition, we report pass@$k$ for both discrete and continuous DLMs using different numbers of samples per prompt (Fig.~\ref{fig:pass_at_k}).

\begin{figure}[!t]
\centering
\vspace{-1mm}
\includegraphics[width=0.99\linewidth]{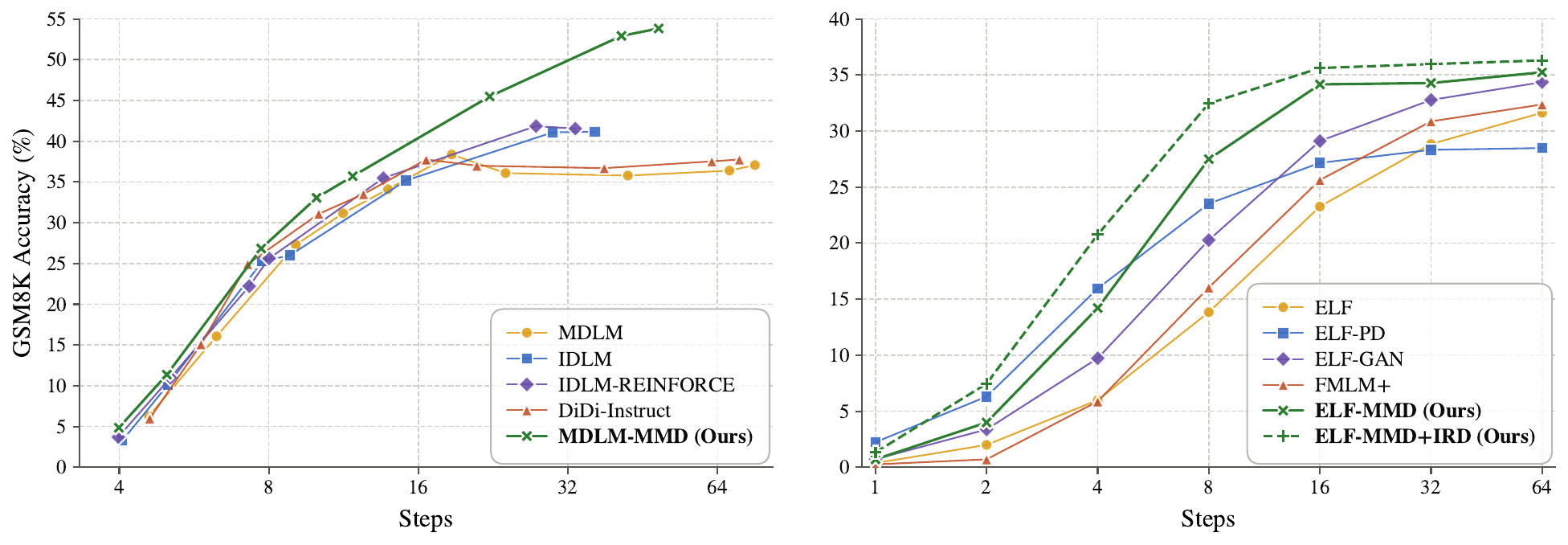}
\vspace{-3mm}
\caption{
    \textbf{GSM8K accuracy vs sampling steps.}
    \textbf{Left:} MDLM-MMD and MDLM-based baselines.
    \textbf{Right:} ELF-MMD and continuous baselines.
    For masked models, we vary the confidence threshold and report the average number of steps per sequence.
}
\label{fig:gsm_results}
\vspace{-4mm}
\end{figure}

\paragraph{Discrete DLMs.}
We initialize MDLM-MMD from an MDLM pretrained on TinyGSM and use frozen MDLM features for the MMD objective. 
We compare against MDLM, IDLM~\citep{li2026idlm}, IDLM-REINFORCE, and DiDi-Instruct~\citep{zheng2026didi}.
Because confidence decoding determines the number of model evaluations adaptively, we sweep the confidence threshold and report the resulting average number of steps per sequence.
As shown in Fig.~\ref{fig:gsm_results}, MDLM-MMD achieves higher accuracy at moderate and high decoding budgets, reaching ${\sim}54\%$ accuracy at ${\sim}49$ steps.

\paragraph{Continuous DLMs.}
Our main comparison uses ELF-B with the final-layer hidden
representations of GPT-2 Small~\citep{radford2019gpt2} as the continuous latent space. 
We additionally evaluate scaling to ELF-M in App.~\ref{app:continuous_scaling}. 
We compare ELF-MMD and ELF-MMD$+$IRD with ELF, ELF-PD~\citep{hu2026elf}, ELF-GAN (see App.~\ref{app:continuous_baselines} for training details), and FMLM$+$~\citep{agarwal2026posterior}, which is initialized from a pretrained masked diffusion model. 
Applying a shifted timestep schedule~\citep{esser2024scaling} improves the accuracy of both ELF and ELF-PD across all evaluated step budgets. 
Thus, we use shift values of 32 and 128, respectively. 


As shown in Fig.~\ref{fig:gsm_results}, ELF-MMD improves accuracy over ELF at every evaluated sampling budget and over ELF-PD at most of them. IRD further improves accuracy at every evaluated budget. At 4 and 8 steps, accuracy rises from \(14.2\%\) to \(20.8\%\) and from \(27.5\%\) to \(32.5\%\), respectively, compared with \(15.9\%\) and \(23.5\%\) for ELF-PD. At 64 steps, ELF-MMD$+$IRD reaches its highest accuracy of \(36.3\%\), compared with \(35.2\%\) for ELF-MMD and \(31.6\%\) for ELF. Exact results are reported in Tab.~\ref{tab:gsm_results}.

We also apply the MMD loss to ELF$^\star$~\citep{yoo2026fixedpoint} and FMLM$+$~\citep{agarwal2026posterior}, yielding ELF$^\star$-MMD and FMLM$+$-MMD. 
Their post-training details and results are provided in App.~\ref{app:additional_exps}.

\vspace{-2mm}
\paragraph{Pass@$k$ results.}
For MDLM-based models, we also evaluate pass@\(k\) using ancestral sampling at temperature 0.
Fig.~\ref{fig:pass_at_k} shows that MDLM-MMD outperforms the baselines at 4–8 sampling steps for \(k \leq 8\).
At 16–32 steps, it also achieves higher pass@\(k\) for most evaluated values of \(k\), with smaller differences at larger sample budgets.
For continuous DLMs, ELF-MMD also improves pass@$k$ at smaller sample budgets, while the gaps narrow as $k$ increases.

\subsection{Scaling to Large Models}
\label{sec:large_models}

To test whether MMD post-training extends to larger models, we start from the released 16B DMax-Math and DMax-Coder checkpoints~\citep{chen2026dmaxaggressiveparalleldecoding}.
DMax is a hybrid masked--uniform diffusion model tuned from LLaDa2.0-Mini~\citep{bie2025llada2} that learns to recover target tokens from both masked inputs and its own token predictions.
We follow the original training implementation and use MMD to train on both masked inputs and the model's own token predictions, yielding DMax-Math-MMD and DMax-Coder-MMD.

We perform 400 training steps that take  ${\sim}13$ minutes for DMax-Math and ${\sim}19$ minutes for DMax-Coder on eight NVIDIA H100 GPUs, corresponding to ${\sim}1.7$ and ${\sim}2.5$ GPU-hours per run, respectively.
These timings cover training only, excluding initialization and evaluation.

We use the released DMax training data, which contain LLaDA-2.0-mini-generated responses to math and code prompts.
For math, we follow the original chain-of-thought evaluation on GSM8K, MATH500~\citep{lightman2024let}, Minerva-Algebra~\citep{hendrycksmath2021}, and ASDIV~\citep{miao2020diverse}.
For code, we report pass@1 on HumanEval-Instruct~\citep{chen2021evaluating} and MBPP-Instruct~\citep{austin2021program}.
Evaluation uses the original dInfer~\citep{dinfer} implementation with Soft Parallel Decoding, a block size of $32$, and a maximum generation length of $2048$ tokens.

We report accuracy and tokens generated per forward (TPF), which measures decoding parallelism.
Tab.~\ref{tab:dmax_scaling} compares these metrics with the original paper's baselines.
We select threshold $0.85$ for DMax-Math-MMD and $0.9$ for DMax-Coder-MMD.
Across the four math benchmarks, DMax-Math-MMD increases TPF by 
$10.3$--$16.5\%$
over the reported DMax-Math operating points while achieving similar or higher accuracy.
On code benchmarks, DMax-Coder-MMD improves accuracy by $2.4$ percentage points on HumanEval-Instruct and $3.8$ points on MBPP-Instruct, while also increasing TPF on both benchmarks.
To show how this trade-off varies with decoding threshold, we provide accuracy--TPF curves in App.~\ref{app:discrete_dmax_protocol}, together with the selected thresholds and aggregation details.

\newcommand{\stk}[2]{\begin{tabular}[c]{@{}c@{}}#1\\[-1pt]{\scriptsize$\pm$#2}\end{tabular}}
\newcommand{\ms}[2]{$#1_{\pm#2}$}

\begin{table}[!t]
\centering
\caption{\textbf{Scaling MMD post-training to 16B DMax.}
    Accuracy ($\%$) and tokens generated per forward (TPF) on math and code benchmarks. 
    Baseline results are taken from the original DMax paper.
}
\vspace{-3mm}
\label{tab:dmax_scaling}
\footnotesize
\setlength{\tabcolsep}{2.5pt}
\renewcommand{\arraystretch}{1.12}
\begin{tabular}{@{}l*{6}{rr}@{}}
\toprule
\multirow{2}{*}{Method}
 & \multicolumn{2}{c}{GSM8K}
 & \multicolumn{2}{c}{MATH500}
 & \multicolumn{2}{c}{\shortstack{Minerva-\\Algebra}}
 & \multicolumn{2}{c}{ASDIV}
 & \multicolumn{2}{c}{\shortstack{HumanEval-\\Instruct}}
 & \multicolumn{2}{c}{\shortstack{MBPP-\\Instruct}} \\
\cmidrule(lr){2-3}\cmidrule(lr){4-5}\cmidrule(lr){6-7}\cmidrule(lr){8-9}\cmidrule(lr){10-11}\cmidrule(lr){12-13}
 & Acc.\,$\uparrow$ & TPF$\uparrow$
 & Acc.\,$\uparrow$ & TPF$\uparrow$
 & Acc.\,$\uparrow$ & TPF$\uparrow$
 & Acc.\,$\uparrow$ & TPF$\uparrow$
 & Acc.\,$\uparrow$ & TPF$\uparrow$
 & Acc.\,$\uparrow$ & TPF$\uparrow$ \\
\midrule
LLaDA-2.0-mini & \textbf{92.6} & 2.04 & \textbf{75.8} & 2.58 & 91.4 & 3.01 & \textbf{92.8} & 2.03 & 84.2 & 4.38 & 80.6 & 2.71 \\
dParallel SFT & 92.3 & 2.79 & \textbf{75.8} & 3.42 & 91.4 & 3.91 & \textbf{93.0} & 2.72 & 76.8 & 5.12 & 74.7 & 3.66 \\
\midrule
DMax-Math & 92.1 & 5.48 & 75.4 & 5.94 & 91.5 & 7.03 & 92.5 & 5.62 & -- & -- & -- & -- \\
\textbf{DMax-Math-MMD} & 92.1 & \textbf{6.15} & \textbf{76.0} & \textbf{6.84} & \textbf{92.1} & \textbf{8.19} & \textbf{92.9} & \textbf{6.20} & -- & -- & -- & -- \\
DMax-Coder & -- & -- & -- & -- & -- & -- & -- & -- & 83.5 & 7.36 & 79.2 & 5.86 \\
\textbf{DMax-Coder-MMD} & -- & -- & -- & -- & -- & -- & -- & -- & \textbf{85.9} & \textbf{8.07} & \textbf{83.0} & \textbf{6.10} \\
\bottomrule
\end{tabular}
\end{table}

\subsection{MMD Loss Ablation}
\label{sec:exp_ablations}

\begin{figure}[!t]
\centering
\includegraphics[width=\linewidth]{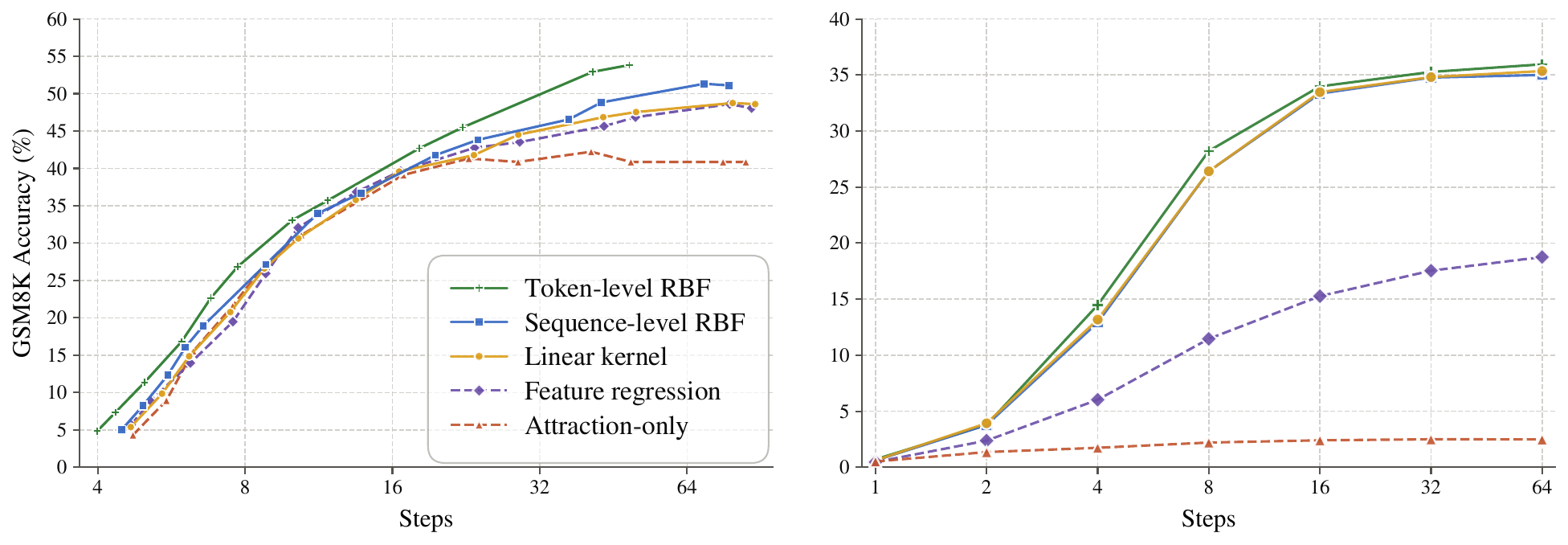}
\vspace{-7mm}
\caption{
    \textbf{MMD loss ablation on GSM8K.} 
    Accuracy vs. sampling steps for MDLM (\textit{left}) and ELF (\textit{right}), post-trained with different feature-based objectives.
    ELF results are obtained without IRD.
}
\label{fig:mmd_loss_comparison}
\end{figure}

Finally, we compare alternative loss formulations on GSM8K (Fig.~\ref{fig:mmd_loss_comparison}), with OWT results in App.~\ref{app:mmd_loss_owt}.
We start with linear-kernel MMD, computed as MSE between generated and reference feature means.
This baseline tests whether matching first moments is sufficient and connects to feature matching for LLMs~\citep{jelassi2026matching}.
Beyond mean matching, we compare sequence-level RBF on mean-pooled embeddings with our token-level RBF.
This aims to address whether retaining token-level interactions improves the MMD training signal.

The attraction-only variant removes the generated--generated repulsive term to test whether similarity to reference features alone suffices.
Finally, a feature regression baseline minimizes MSE between the full intermediate feature tensors of sampled and reference sequences at corresponding positions.

On GSM8K, token-level RBF gives the strongest accuracy--computation trade-off, with larger gains for discrete models and more modest improvements for continuous models.
Attraction-only and feature regression losses yield lower accuracy, particularly for continuous models.
The OWT results in Fig.~\ref{fig:owt_loss_comparison} similarly favor token-level RBF for both discrete and continuous models in most setups. 
These results support our choice of token-level RBF.

\section{Discussion}
Our results show that MMD on pretrained DLM features can improve generation in both discrete and continuous DLMs.
These improvements are obtained with a simple and efficient training procedure that requires neither full sampling trajectories nor jointly trained auxiliary models.
Nevertheless, performance depends on choices such as the RBF bandwidth and the representation space used for distribution matching.
This makes representation design a natural direction for further exploration.
We use features from a single layer on clean inputs, but combining features across layers or noise levels, or using an ensemble of extractors, could provide complementary information.

Alongside representation design, future work could explore combining MMD with other training objectives for discrete DLMs. 
Such combinations may offer further benefits in practice, as suggested by the gains from applying IRD after ELF-MMD post-training in the continuous setting.

\bibliography{iclr2027_bibtex}
\bibliographystyle{iclr2027_conference}

\clearpage
\appendix
\section{Related Work}
\label{app:related_work}

\paragraph{Diffusion language models.}
Diffusion language modeling can be formulated over discrete tokens or continuous representations of text.
Masked models such as MDLM~\citep{sahoo2024simple} and MD4~\citep{shi2024simplified} learn to recover missing tokens, a formulation that also underlies many large discrete DLMs~\citep{nie2025llada,bie2025llada2,fu2026nemotronlabsdiffusion}.
Uniform DLMs instead replace tokens with uniformly sampled vocabulary items during corruption~\citep{hoogeboom2021argmaxflowsmultinomialdiffusion,schiff2025simpleguidancemechanismsdiscrete,sahoo2025the}.
Their reverse process can revise token identities throughout denoising, as in DiffusionGemma~\citep{diffusiongemmateam2026diffusiongemma}.
Continuous models denoise token embeddings~\citep{li2022diffusionlm, chen2026langflow, deschenaux2026language}, encoder latents~\citep{lovelace2023ld4lg,meshchaninov2025cosmos, meshchaninov2026ldlm, hu2026elf}, or noisy one-hot representations~\citep{lee2026fmlm, agarwal2026posterior}.
Our continuous realization builds on ELF~\citep{hu2026elf}, which learns a flow in a frozen contextual embedding space. 

\paragraph{One- and few-step continuous generation.}
One way to reduce sampling cost is to train a model to reproduce several denoising steps at once.
Direct and progressive distillation~\citep{luhman2021knowledge,salimans2022progressive} learn from teacher-generated targets, while consistency models~\citep{song2023consistency} and flow-map matching~\citep{boffi2024flowmap} provide related ways to learn large transitions through the generative process.
For language, ELF applies progressive distillation, and FMLM~\citep{lee2026fmlm} learns the flow map of a continuous language model.
FMLM+~\citep{agarwal2026posterior} extends FMLM with masking-style noise schedules and posterior-guided refinement.
FMLM$^\star$~\citep{yoo2026fixedpoint} combines fixed-point and flow-map distillation for self-conditioned models.
DLM-One~\citep{chen2025dlmone} takes a distribution-matching approach based on score distillation and adversarial regularization.
Our MMD stage trains a latent generator through comparisons with reference samples in a fixed feature space.
We use self-conditioning to refine its predictions at a fixed diffusion time and optionally distill these refinement steps afterward.

\paragraph{Few-step discrete DLMs.}
SDTT~\citep{deschenaux2024sdtt} accelerates discrete DLMs by training a student to match token distributions collected across several teacher denoising steps.
Inspired by consistency models~\citep{song2023consistency}, CDLM~\citep{kim2025cdlm} combines teacher-logit distillation with consistency regularization across discrete denoising states.
Other approaches construct distribution-matching objectives using learned auxiliary models.
IDLM~\citep{li2026idlm} and Discrete Moment Matching Distillation (D-MMD)~\citep{hoogeboom2026dmmd} train an auxiliary denoiser and update the generator by differentiating through soft token predictions.
DiDi-Instruct~\citep{zheng2026didi} instead uses policy gradients with a learned density-ratio discriminator.
Our discrete formulation also uses policy gradients, with rewards computed directly from MMD in frozen features.


\paragraph{Distribution matching in representation space.}
MMD has long been used to train implicit generators from samples~\citep{li2015gmmn}.
Its effectiveness depends on the representation in which samples are compared.
MMD-GAN~\citep{li2017mmdgan} learns this representation adversarially, whereas GFMN~\citep{Santos_2019_ICCV} matches feature statistics from frozen pretrained networks.
For diffusion models, MMD-DDM~\citep{aiello2024fast} fine-tunes samplers for a fixed step budget.
SwD~\citep{starodubcev2026swd} brings these ideas together through patch-level MMD on pretrained diffusion features, directly motivating our use of contextual token features.
Inductive Moment Matching~\citep{zhou2025imm} uses MMD to enforce distributional consistency across sampling intervals.
Related objectives include Representation Fr\'echet Loss~\citep{yang2026frechet}, which compares feature means and covariances, and Drifting Models~\citep{deng2026drifting, oba2026tokendrift}, which regress to targets defined by an attraction--repulsion field.

\paragraph{Feature matching for language.}
Frozen-feature matching has also been explored directly for text generation.
SeqGFMN~\citep{padhi2020learning} matches position-wise means and variances of token features and differentiates through soft token outputs.
More recently, Energy-Based Fine-Tuning (EBFT)~\citep{jelassi2026matching} fine-tunes autoregressive models using frozen features of short rollouts and policy-gradient updates.
Its basic, unwhitened objective corresponds to conditional linear-kernel MMD, with whitening and reward modifications in the practical training recipe.
EBFT represents each rollout by a single feature vector.
We instead compare distributions of token features using an RBF kernel, applying the objective to parallel discrete predictions or continuous latent samples.



\section{Continuous DLMs Training Details}

\subsection{Iterative Refinement Distillation}
\label{sec:ird}

While bootstrapping aims to mitigate the training--inference mismatch for iterative refinement, we also aim to further reduce the number of refinement steps.
To this end, inspired by~\citep{yoo2026fixedpoint}, we employ \emph{Iterative Refinement Distillation} (IRD), which adapts a simple Knowledge Distillation (KD) approach~\citep{luhman2021knowledge}, i.e., distills multiple refinement iterations into a single prediction.

Specifically, we freeze the trained ELF-MMD generator as a teacher $G_{\bar{\theta}}$ and generate a fixed $K$-step trajectory using the same noise at every iteration:
\begin{equation}
\begin{aligned}
    \bar{\mathbf{x}}^{(0)}=\mathbf{0},
    \qquad
    \bar{\mathbf{x}}^{(k)}=
    G_{\bar{\theta}}\!\left(
        \mathbf{z},t{=}0;
        \mathrm{SC}{=}\bar{\mathbf{x}}^{(k-1)}
    \right),
    \qquad
    k=1,\ldots,K,
    \qquad
    \mathbf{z}\sim
    \mathcal{N}(\mathbf{0},\mathbf{I}).
\end{aligned}
\label{eq:ird_teacher_rollout}
\end{equation}

Using the same noise $\mathbf{z}$, we train the student with $SC{=}0$ to predict the final teacher output:
\begin{equation}
    \mathcal{L}_{\mathrm{IRD}}(\theta)
    =
    \mathbb{E}_{\mathbf{z}\sim\mathcal{N}(\mathbf{0},\mathbf{I})}
    \left[
        \left\|
            G_\theta\!\left(
                \mathbf{z},t{=}0;
                \mathrm{SC}{=}\mathbf{0}
            \right)
            -
            \bar{\mathbf{x}}^{(K)}
        \right\|_2^2
    \right].
\label{eq:ird_loss}
\end{equation}
Thus, the student is trained to reproduce the output obtained after $K$ teacher refinement steps.
The pseudocode is presented in \alg{ours_ird}.
We also experimented with more sophisticated approaches, e.g., consistency distillation~\citep{song2023consistency} or the method used in FMLM$^\star$~\citep{yoo2026fixedpoint} but did not observe noticeable improvements. 

IRD is an optional trajectory-supervised stage after MMD training.
The MMD stage itself does not use paired teacher trajectories.
At inference, the distilled generator can again be used with iterative self-conditioning.

\subsection{Decoder Fine-tuning}
\label{app:decoder_finetune}

Alongside MMD post-training, we fine-tune the model in decoding mode to reconstruct the original tokens from perturbed data latents using a cross-entropy loss, following ELF~\citep{hu2026elf}. We increase the global batch size by $\sim20\%$ to include additional samples for decoder fine-tuning.

\section{Ablation study} 
\label{app:additional_abl}

Here, we evaluate how the design choices described in Section~\ref{sec:method} affect performance.
For ELF-MMD, we omit IRD from these experiments. 




    


\subsection{MMD Loss Ablation on OWT}
\label{app:mmd_loss_owt}

\begin{figure}[t]
    \centering
    \vspace{-4mm}
    \includegraphics[width=\linewidth]{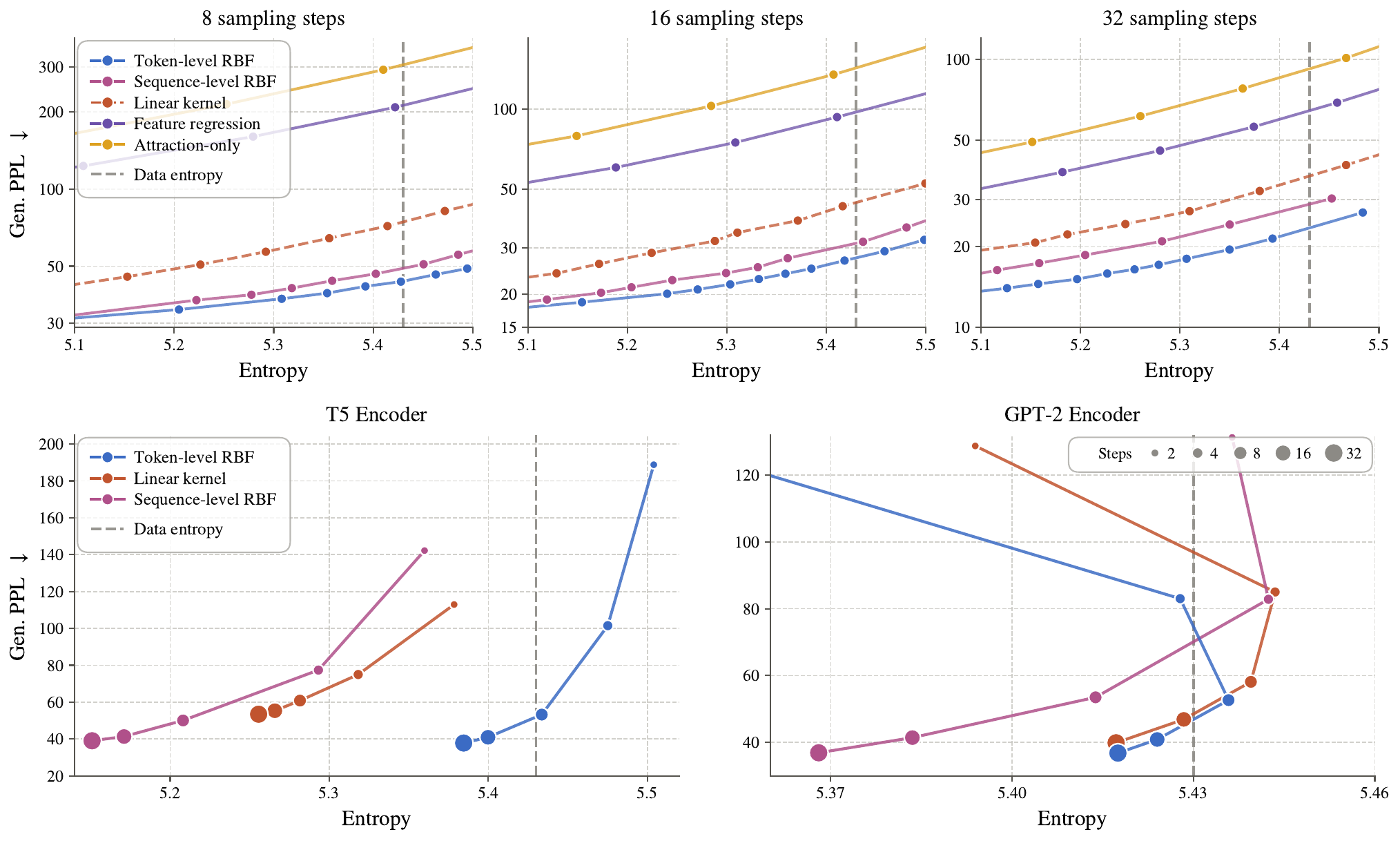}
    \vspace{-7mm}
    \caption{
        \textbf{MMD loss ablation on OpenWebText.} 
        Attraction-only and feature regression losses are shown only for MDLM, as both perform poorly on ELF in the OWT experiments.
    }
    \label{fig:owt_loss_comparison}
\end{figure}

Fig.~\ref{fig:owt_loss_comparison} extends the comparisons in Section~\ref{sec:exp_ablations} to unconditional generation.
We compare linear-kernel feature matching and sequence and token level RBF formulations, attraction-only and feature regression losses.

\paragraph{Discrete DLMs.}
At all three sampling budgets, token-level RBF yields the lowest generative perplexity at matched entropy, followed by sequence-level RBF and the linear kernel.
Attraction-only and feature regression losses substantially worsen this trade-off, reinforcing the GSM8K evidence for retaining token features and the generated--generated term.

\paragraph{Continuous DLMs.}
The effect of kernel choice and feature aggregation depends on the latent space. 
With T5 latents, token-level RBF achieves a better perplexity–entropy trade-off than both linear and sequence-level RBF kernels. 
With GPT-2 latents, it provides a modest improvement over the linear kernel. 
We omit the attraction-only and feature-regression baselines because both lead to collapse.


\subsection{Representation Spaces}
\label{app:encoders}

\begin{figure}[!t]
\centering
\vspace{-3mm}
\includegraphics[width=\linewidth]{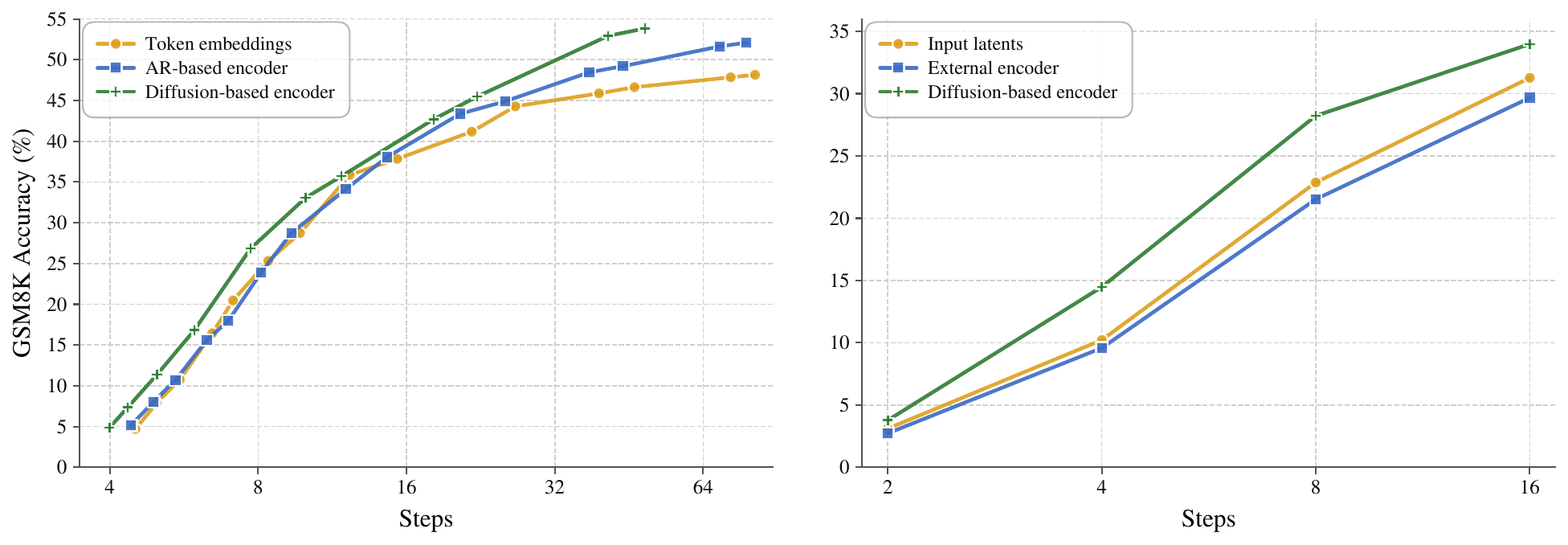}
\vspace{-7mm}
\caption{
    \textbf{Representation-space ablation on GSM8K.}
    Accuracy vs sampling steps for MDLM (\textit{left}) and ELF (\textit{right}), comparing input representations, external encoders, and pretrained DLM features.
}
\label{fig:encoders_abl}
\end{figure}

Here, we address whether intermediate DLM features improve MMD training over input representations, and whether separately pretrained encoders offer a suitable alternative. Fig.~\ref{fig:encoders_abl} compares these choices on GSM8K using MDLM and ELF, with setups tuned separately for each representation type.


\subsubsection{Input Representations}

\paragraph{Discrete DLMs.} We compute MMD directly on token embeddings from the frozen MDLM's vocabulary embedding matrix. Unlike intermediate hidden states, these embeddings depend only on token identity and do not capture the surrounding context.

\paragraph{Continuous DLMs.} We compute MMD directly in the latent space of the pretrained text encoder, using the identity feature map \(\phi(x)=x\). These latents already encode contextual information, allowing distribution matching without an additional feature extractor.

\subsubsection{External Encoder}

\paragraph{Discrete DLMs.}
We use a frozen autoregressive language model pretrained on the same dataset. We extract contextual token features from an intermediate layer for both generated and reference sequences and compute MMD over these features.

\paragraph{Continuous DLMs.}
Inspired by the latent-space MAE encoder in Drifting Models~\citep{deng2026drifting}, we train a separate Transformer on real ELF latents using masked-token cross-entropy rather than latent reconstruction.

\paragraph{Results.}

Diffusion-based features achieve the highest accuracy for both discrete and continuous DLMs. They outperform token embeddings and autoregressive features in the discrete setting, as well as input latents and external-encoder features in the continuous setting. 



\begin{figure}[!t]
\centering
\includegraphics[width=\linewidth]{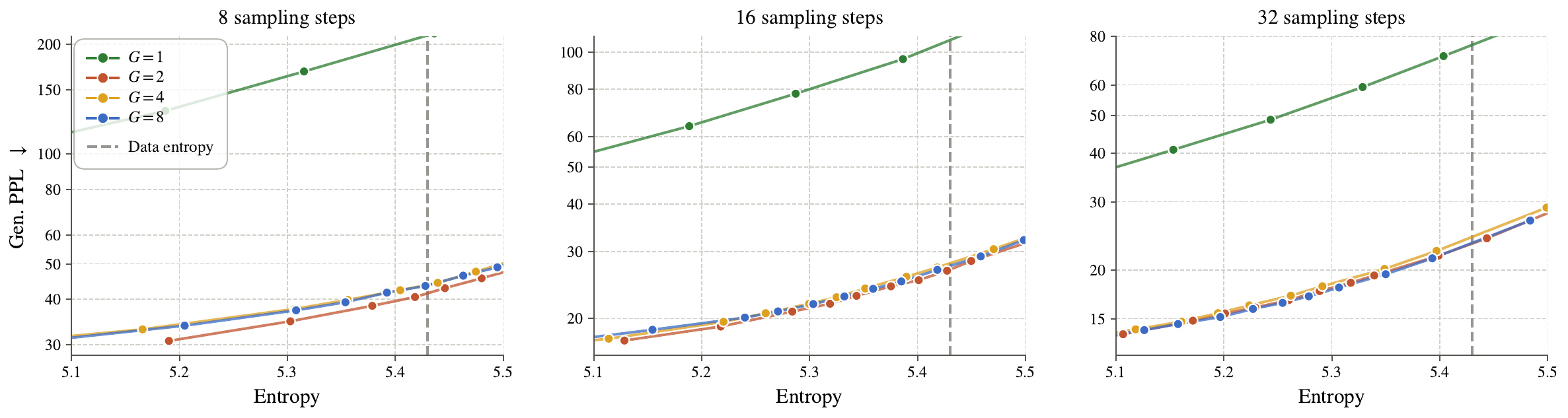}
\caption{\textbf{Ablation on policy-gradient group size on OWT.} Generative perplexity--entropy trade-offs at 8, 16, and 32 sampling steps. Increasing $G$ from $1$ to $2$ yields the largest improvement.}
\label{fig:discrete_group_size}
\end{figure}

\subsection{Policy-gradient Group Size}

\begin{wrapfigure}{r}{0.49\textwidth}
\vspace{-10mm}
\centering
\includegraphics[width=\linewidth]{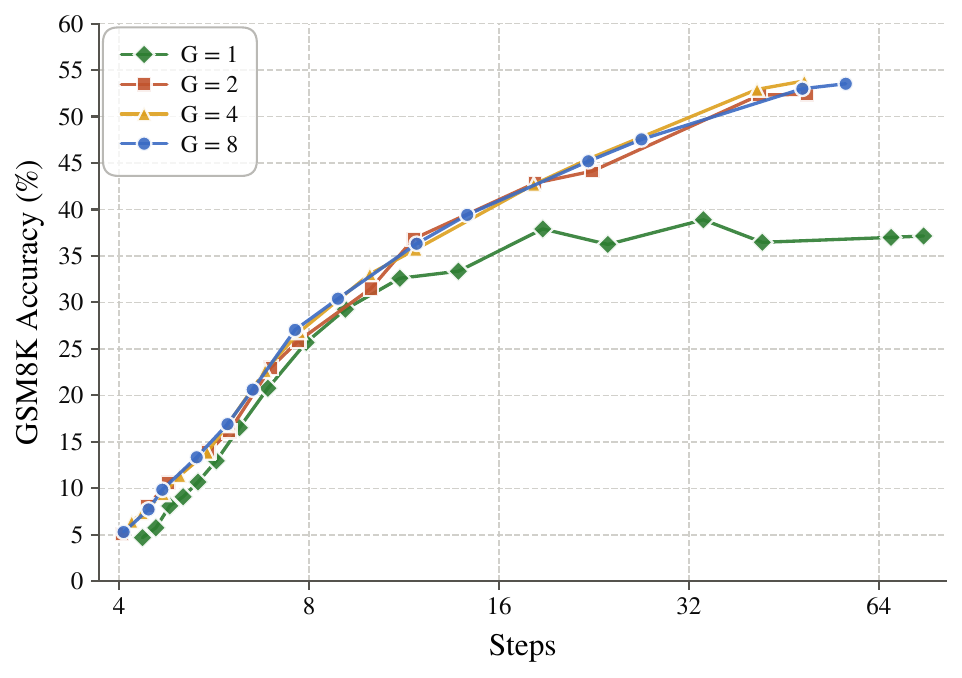}
\vspace{-8mm}
\caption{\textbf{Ablation on policy-gradient group size on GSM8K.} Accuracy vs. average number of sampling steps.}
\label{fig:discrete_group_size_gsm}
\end{wrapfigure}

We vary the group size used for policy-gradient
estimation (Section~\ref{sec:discrete}), $G\in\{1,2,4,8\}$ on OpenWebText and GSM8K. 
Fig.~\ref{fig:discrete_group_size} shows the generative perplexity--entropy trade-offs on OWT at 8, 16, and 32 sampling steps, while Fig.~\ref{fig:discrete_group_size_gsm} shows answer accuracy versus the average number of sampling steps on GSM8K.
The largest improvement occurs when increasing $G$ from $1$ to $2$, which enables the leave-one-out baseline. Further increases yield negligible gains: $G\in\{2,4,8\}$ produce similar trade-offs. We therefore adopt $G=4$ as a shared default and
keep it fixed across all other discrete MMD
post-training experiments on OWT and GSM8K.

\par
\makeatletter
\ifnum\c@WF@wrappedlines>\@ne
  \vskip\dimexpr\c@WF@wrappedlines\baselineskip-\baselineskip\relax
\fi
\WFclear
\makeatother

\subsection{Model-Size Scaling}
\label{app:continuous_scaling}

\begin{wrapfigure}{r}{0.49\textwidth}
\vspace{-10mm}
\centering
\includegraphics[width=\linewidth]{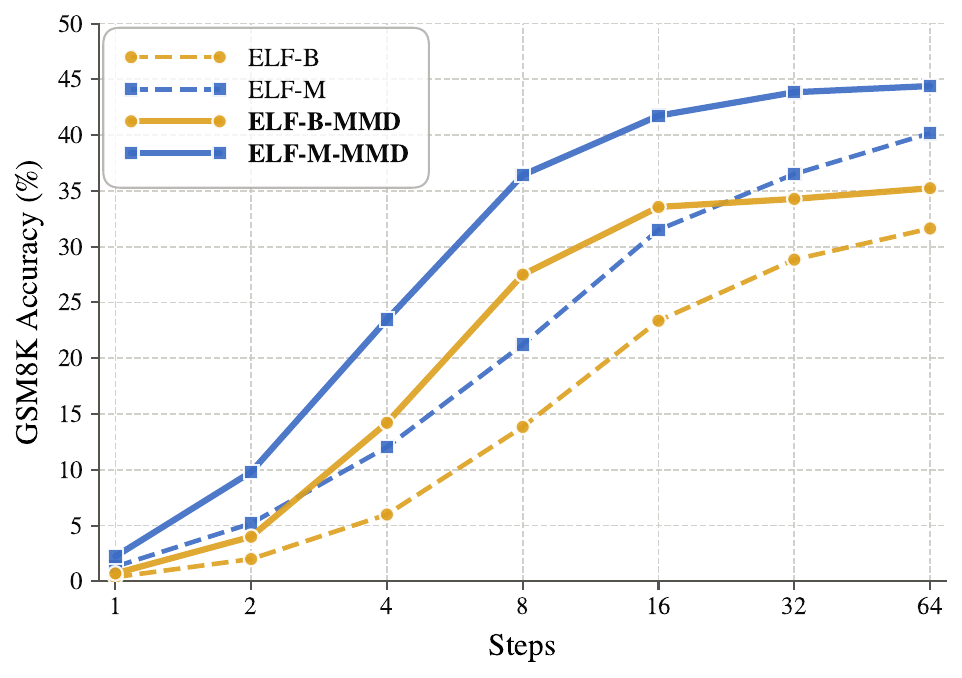}
\vspace{-8mm}
\caption{\textbf{Model-size scaling on GSM8K.} ELF-B and ELF-M before (dashed) and after (solid) MMD post-training.}
\label{fig:gsm_scaling_ablation}
\end{wrapfigure}

To assess scaling to larger continuous DLMs, we train ELF-M on TinyGSM and post-train it with MMD. We omit IRD in this comparison. We use the same GPT-2 Small latent space as in the ELF-B experiments and evaluate both model sizes on GSM8K. As shown in Fig.~\ref{fig:gsm_scaling_ablation}, increasing model size improves accuracy both before and after MMD post-training across all evaluated sampling budgets. ELF-M-MMD also outperforms the original ELF-M at every budget, indicating that the benefits of our method persist when scaling from ELF-B to ELF-M. The smaller ELF-B-MMD achieves higher accuracy than ELF-M at $4$--$16$ steps and similar accuracy at $32$ steps.

\par
\makeatletter
\ifnum\c@WF@wrappedlines>\@ne
  \vskip\dimexpr\c@WF@wrappedlines\baselineskip-\baselineskip\relax
\fi
\WFclear
\makeatother

\subsection{Bootstrapping}

\begin{figure}[!t]
    \centering
    \includegraphics[width=\linewidth]{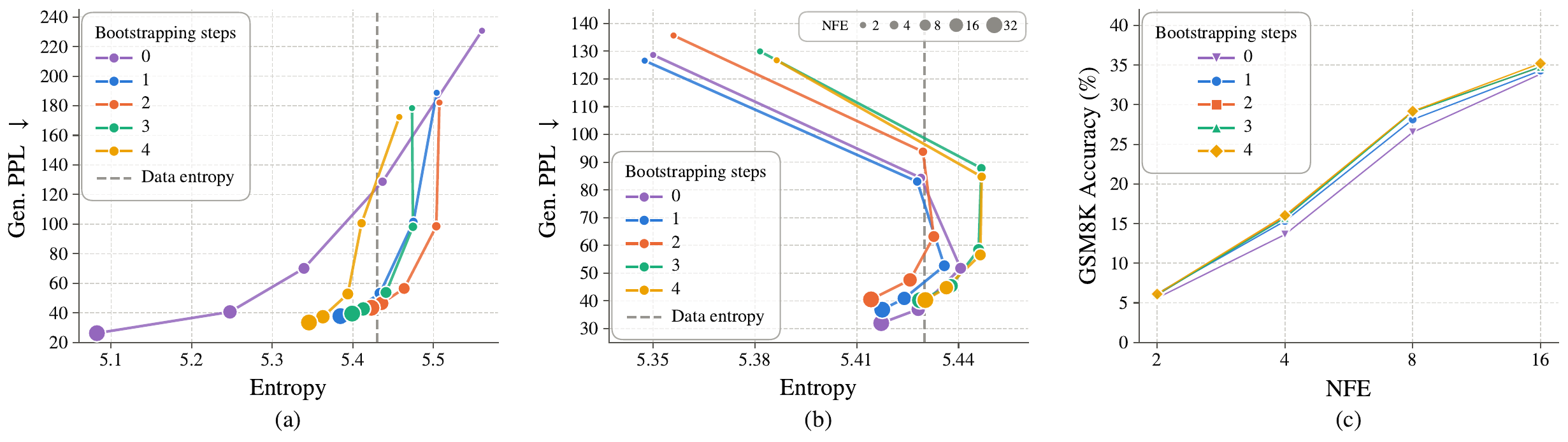}
    \caption{\textbf{Ablation on bootstrap training.} We vary the number of bootstrap steps on OWT with T5 (a) and GPT-2 (b) encoders, and on GSM8K (c). A single bootstrap step improves the trade-off with T5 and accuracy on GSM8K, while results with GPT-2 are mixed.}
    \label{fig:bootstrap_ablation}
\end{figure}

As discussed in Section~\ref{sec:bootstrap}, bootstrapping helps reduce the mismatch between training and inference. We vary the number of bootstrap steps during generator training, including a baseline without bootstrapping. As shown in Fig.~\ref{fig:bootstrap_ablation}, a single bootstrap step already provides performance gains, while additional steps yield limited or inconsistent improvements. Since each additional step increases training costs, a small number of bootstrap steps offers a practical balance between generation quality and computational overhead.

\section{Additional DMax results}
\label{app:discrete_dmax_protocol}



\begin{figure}[!t]
\centering
\includegraphics[width=\linewidth]{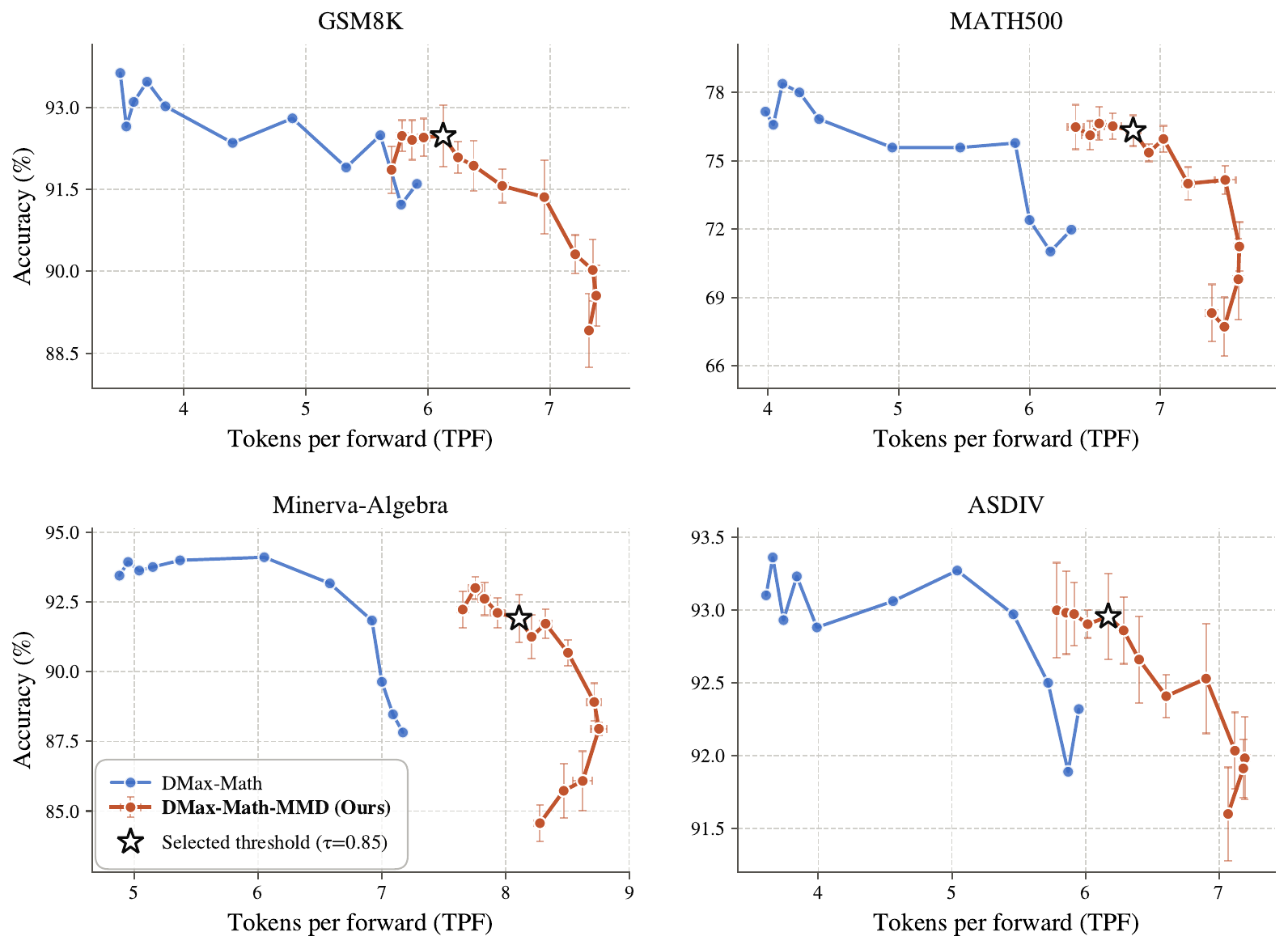}
\caption{\textbf{Accuracy--TPF trade-offs on math benchmarks.}
DMax-Math and DMax-Math-MMD are evaluated at eleven decoding thresholds.
MMD points average five training runs with different seeds. 
Error bars show $\pm1$ sample standard deviation in both coordinates.
Lines connect points in threshold order, and stars mark the operating points in Tab.~\ref{tab:dmax_scaling}.}
\label{fig:dmax_math_tradeoff}
\end{figure}

\begin{figure}[!t]
\centering
\includegraphics[width=\linewidth]{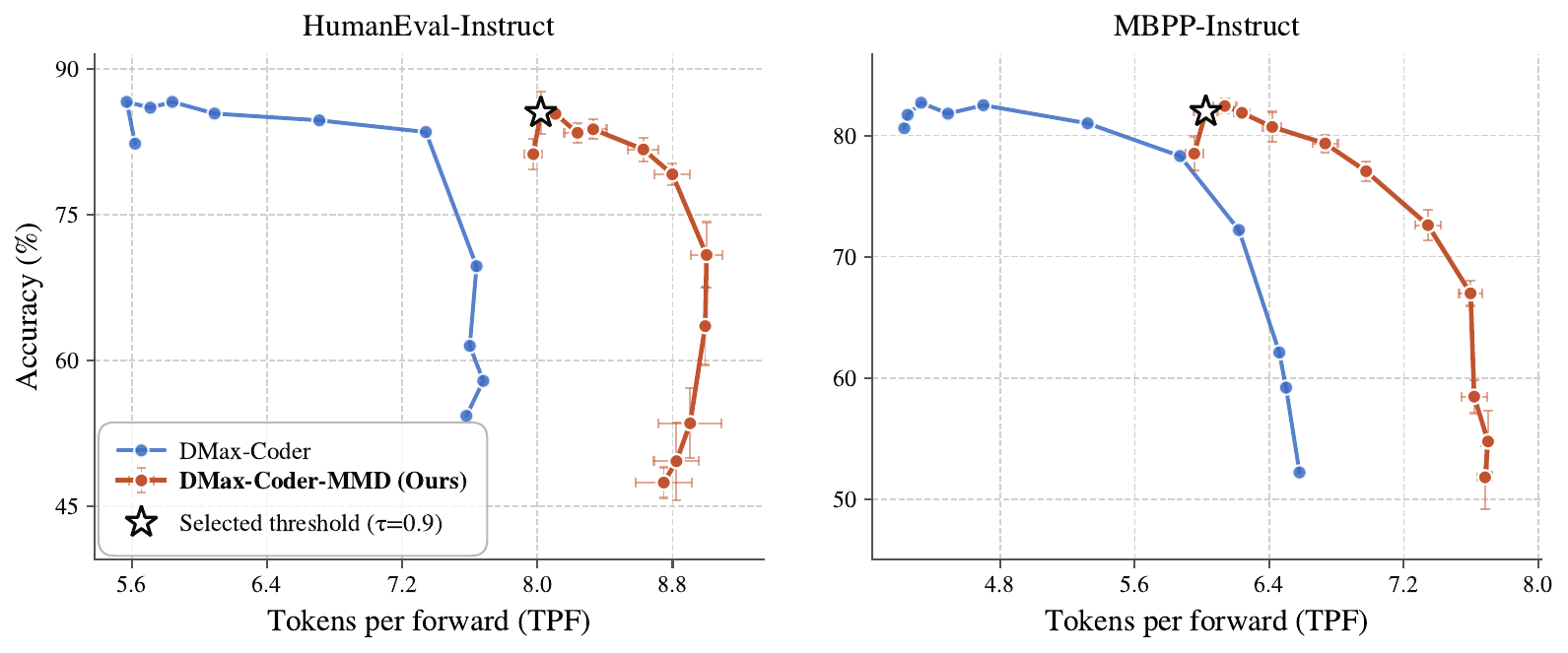}
\caption{\textbf{Accuracy--TPF trade-offs on code benchmarks.}
DMax-Coder and DMax-Coder-MMD are evaluated at eleven decoding thresholds.
Accuracy denotes pass@1.
MMD points average five training runs with different seeds. 
Error bars show $\pm1$ sample standard deviation in both coordinates.
Lines connect points in threshold order, and stars mark the operating points in Tab.~\ref{tab:dmax_scaling}.}
\label{fig:dmax_code_tradeoff}
\end{figure}

We provide additional results for DMax-Math-MMD (Fig.~\ref{fig:dmax_math_tradeoff}) and DMax-Coder-MMD (Fig.~\ref{fig:dmax_code_tradeoff}), examining the accuracy--TPF trade-off across decoding thresholds on math and code benchmarks.
The uncertainties are sample standard deviations across the five independent training runs.
The baseline curves use our evaluations of the released DMax checkpoints, with one evaluation per threshold.
The main table retains the baseline values reported in the DMax paper since we could accurately reproduce the original DMax results using the official implementation.
The evaluated thresholds are $[0.25, 0.35, 0.45, 0.55, 0.65, 0.75, 0.8, 0.85, 0.875, 0.89, 0.9, 0.91]$.
Stars mark the selected thresholds for the MMD-tuned models.
These marked results differ slightly from those in Tab.~\ref{tab:dmax_scaling} due to averaging over five training runs.


\section{Baselines Details}

\vspace{-2mm}

\subsection{Masked DLM Baselines}
\label{app:masked_baselines}

\paragraph{IDLM.}
We reproduce IDLM~\citep{li2026idlm} using the authors' released
code for both OWT and TinyGSM, following their MDLM-based
training procedure.

\paragraph{IDLM-REINFORCE.}
We build IDLM-REINFORCE on the same implementation, replacing the original student update with the policy-gradient optimization in Eq.~\ref{eq:discrete_surrogate}.
The IDLM log-ratio objective supplies the reward for sampled discrete completions, while the teacher and auxiliary-denoiser training procedure are retained.

\paragraph{DiDi-Instruct.}
For OWT, we evaluate the publicly released
DiDi-Instruct checkpoint~\citep{zheng2026didi} using the authors'
public inference code, without additional training.
For TinyGSM, we adapt their OWT training code to conditional
generation, following the approach used by IDLM to extend
MDLM distillation to TinyGSM.

\textbf{D-MMD}~\citep{hoogeboom2026dmmd} provides neither an implementation nor model checkpoints. We therefore report its OWT results from the original paper.

\textbf{TokenDrift}~\citep{oba2026tokendrift} reports results only on OWT and likewise provides neither an implementation nor model checkpoints. We therefore use the results reported in the original paper.

\vspace{-2mm}

\subsection{Continuous DLM Baselines}
\label{app:continuous_baselines}

\paragraph{ELF.} For OWT, we use the authors’ released code and checkpoints for both T5 and GPT-2 encoders~\citep{hu2026elf, yoo2026fixedpoint}. For TinyGSM, we reuse the code provided for ELF’s conditional-generation experiments on other datasets.

\paragraph{ELF-PD.}

For OWT, we use the released code and checkpoint for the T5-based model~\citep{hu2026elf} and train the GPT-2-based variant using the same implementation. For TinyGSM, we follow the same distillation setup, adding one more round for 64 steps.

\paragraph{ELF$^\star$.}

For OWT, we use the released code and checkpoint for the GPT-2-based model~\citep{yoo2026fixedpoint} and train the T5-based variant using the same implementation. For TinyGSM, we adapt this code to conditional generation and additionally use classifier-free guidance (CFG) during training. We find that the resulting model still requires self-conditioning, so we use ELF-style sampling with self-conditioning for the TinyGSM checkpoint only.

\paragraph{ELF-GAN.}

We initialize ELF-GAN from pretrained ELF-B checkpoints and adversarially post-train it on OWT and TinyGSM. A four-layer MLP discriminator operates on token features extracted by a frozen ELF. We use logistic discriminator and non-saturating generator losses, averaged over non-prompt positions, with five discriminator updates per generator update. We retain decoder fine-tuning (App.~\ref{app:decoder_finetune}) and use the same generator optimizer and learning rate as ELF-MMD. A sequence-level variant that mean-pools features over non-prompt positions performs similarly to the token-level objective.




\section{Extending MMD to Other Continuous DLMs}
\label{app:additional_exps}

\subsection{ELF$^\star$-MMD}
\label{app:elf_star_mmd}

We apply our MMD objective to ELF$^\star$ following the same training procedure as ELF-MMD. Features are extracted using a frozen copy of ELF$^\star$. Since the TinyGSM model still requires self-conditioning, we retain it during inference, following the same iterative refinement sampling procedure as ELF-MMD. On GSM8K, ELF$^\star$-MMD achieves higher accuracy than ELF$^\star$ at every evaluated sampling budget (Tab.~\ref{tab:gsm_results}).

\subsection{FMLM$+$-MMD}
\label{app:fmlm_plus_mmd}

We also apply our objective to FMLM$+$, which extends flow-map language modeling with posterior refinement. Features are extracted using a frozen copy of FMLM$+$. Training runs for $1{,}000$ steps with a learning rate of \(5{\times}10^{-6}\), using otherwise the same settings as ELF-MMD. We evaluate FMLM$+$-MMD using the same protocol as FMLM$+$, with the reported number of steps denoting the number of refinement steps. The resulting FMLM$+$-MMD achieves higher accuracy than FMLM$+$ on GSM8K at every evaluated sampling budget (Tab.~\ref{tab:gsm_results}).

\section{Hyperparameters}

Tabs.~\ref{tab:mdlm_hyperparameters}, \ref{tab:elf_hyperparameters}, and~\ref{tab:dmax_hyperparameters} summarize the hyperparameters for MDLM-MMD, ELF-MMD, and DMax-MMD, respectively, covering MMD settings, feature extraction, and optimization. We provide these details to document our experimental setup and help others reproduce our results.

\begin{figure}[!t]
    \centering
    \includegraphics[width=\linewidth]{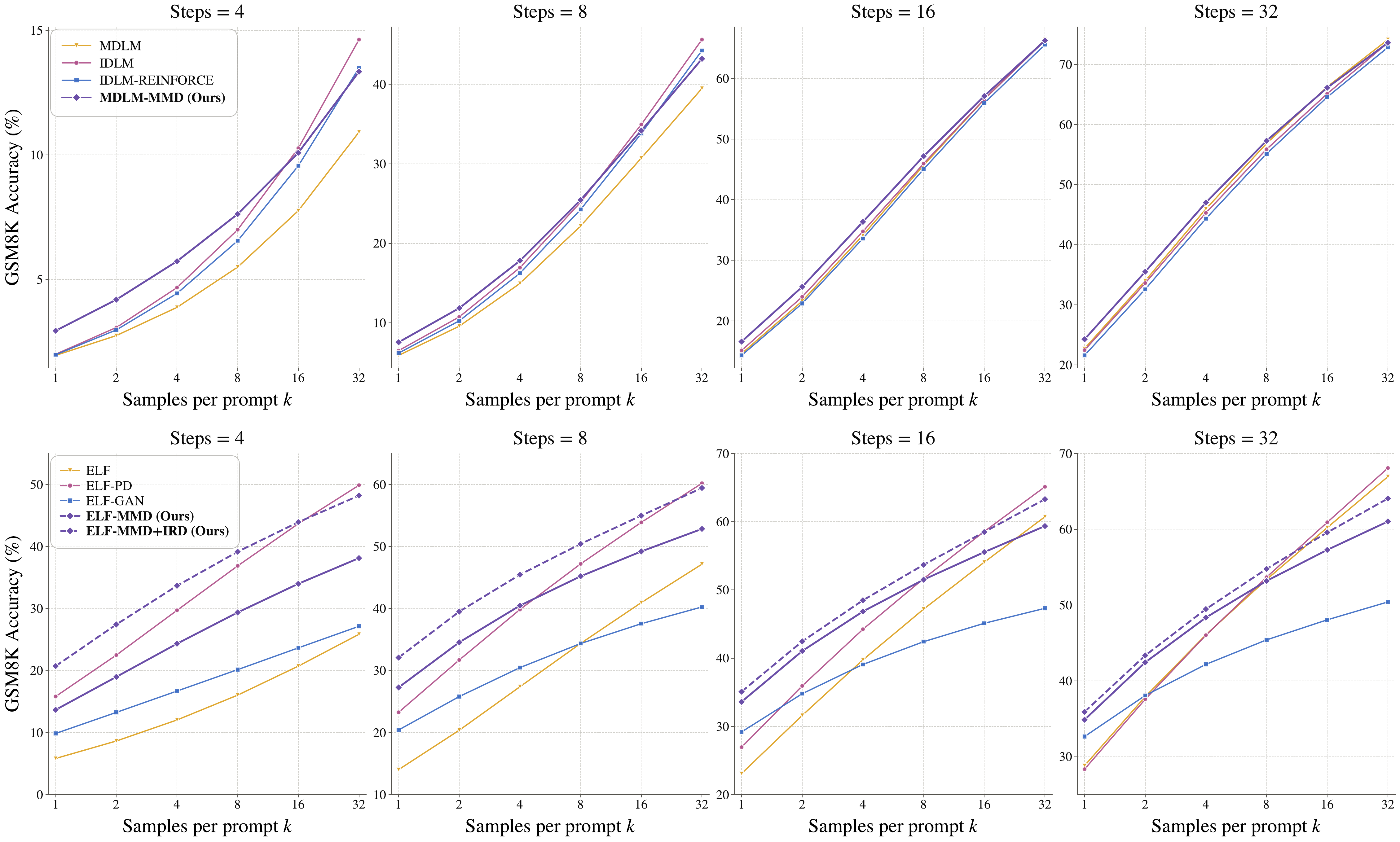}    
    \caption{\textbf{Pass@$k$ on GSM8K.}
    We report pass@$k$ accuracy for different numbers of samples
    per prompt and sampling budgets of 4-32 steps for both discrete and continuous setups. 
    }
    \label{fig:pass_at_k}
\end{figure}


\begin{table}[t!]
\centering
\caption{\textbf{MDLM results on OpenWebText.}
    We report generative perplexity (gPPL, $\downarrow$) and unigram entropy (Ent.) across sampling budgets. 
    For each method and budget, we select the evaluated point whose entropy is closest to the data entropy ($H=5.43$). The results are averaged by 5 seeds.
    \graycell{Gray} marks results from the original papers with entropy substantially different from the data entropy.
}
\label{tab:discrete_owt_resutls}
\setlength{\tabcolsep}{3.5pt}
\small
\begin{tabular}{l | cc cc cc}
\toprule
\multicolumn{1}{c|}{Steps}
& \multicolumn{2}{c}{8}
& \multicolumn{2}{c}{16}
& \multicolumn{2}{c}{32} \\
\cmidrule(lr){2-3}
\cmidrule(lr){4-5}
\cmidrule(lr){6-7}
\multicolumn{1}{c|}{Method}
& gPPL & Ent. & gPPL & Ent. & gPPL & Ent. \\
\midrule
\graycell{Masked D-MMD}
& \graycell{---} & \graycell{---}
& \graycell{17.2} & \graycell{5.00}
& \graycell{19.4} & \graycell{5.05} \\
\graycell{Masked D-MMD ($p=1.0$)}
& \graycell{---} & \graycell{---}
& \graycell{67.7} & \graycell{5.57}
& \graycell{72.1} & \graycell{5.57} \\
\graycell{TokenDrift}
& \graycell{\stk{55.04}{1.96}} & \graycell{\stk{5.30}{0.02}}
& \graycell{\stk{27.35}{0.20}} & \graycell{\stk{5.16}{0.02}}
& \graycell{---} & \graycell{---} \\
\midrule
MDLM
& \stk{282.47}{2.52} & \stk{5.45}{0.00}
& \stk{127.31}{1.87} & \stk{5.44}{0.02}
& \stk{73.63}{1.50}  & \stk{5.42}{0.01} \\

DiDi-Instruct
& \stk{149.01}{5.00} & \stk{5.44}{0.03}
& \stk{104.14}{5.99} & \stk{5.52}{0.01}
& \stk{57.42}{2.54}  & \stk{5.41}{0.02} \\

IDLM
& \stk{56.20}{0.68} & \stk{5.44}{0.01}
& \stk{33.77}{0.58} & \stk{5.44}{0.01}
& \stk{29.30}{0.58} & \stk{5.44}{0.02} \\

IDLM-REINFORCE
& \stk{52.93}{0.36} & \stk{5.41}{0.01}
& \stk{36.00}{0.55} & \stk{5.43}{0.01}
& \stk{34.30}{0.81} & \stk{5.47}{0.01} \\

\midrule

\textbf{MDLM-MMD}
& \stk{\textbf{43.50}}{0.86} & \stk{5.43}{0.01}
& \stk{\textbf{26.84}}{0.48} & \stk{5.42}{0.01}
& \stk{\textbf{21.46}}{0.18} & \stk{5.39}{0.01} \\
\bottomrule
\end{tabular}
\end{table}

\begin{table}[t!]
\centering
\caption{\textbf{Continuous DLMs on OpenWebText.} For each number of sampling steps we report generative perplexity (gPPL, $\downarrow$) and unigram entropy (Ent.). \textbf{Bold} highlights the lowest perplexity at comparable entropy. The results are averaged by 5 seeds.
}
\label{tab:owt_resutls}
\setlength{\tabcolsep}{3.5pt}
\small
\begin{tabular}{l l | cc cc cc cc cc}
\toprule
\multicolumn{2}{c|}{Steps}
& \multicolumn{2}{c}{2} & \multicolumn{2}{c}{4}
& \multicolumn{2}{c}{8} & \multicolumn{2}{c}{16} & \multicolumn{2}{c}{32} \\
\cmidrule(lr){3-4} \cmidrule(lr){5-6} \cmidrule(lr){7-8}
\cmidrule(lr){9-10} \cmidrule(lr){11-12}
Encoder & \multicolumn{1}{c|}{Method}
&  gPPL & Ent. & gPPL & Ent.
& gPPL & Ent. & gPPL & Ent. & gPPL & Ent. \\
\midrule
\multirow{11}{*}[-2.4em]{T5}    
& ELF            & \stk{34.02}{6.95} & \stk{4.38}{0.30}  & \stk{116.38}{6.61} & \stk{5.48}{0.05} & \stk{91.77}{2.70} & \stk{5.38}{0.01} & \stk{57.77}{1.42} & \stk{5.36}{0.01} & \stk{45.13}{1.06} & \stk{5.30}{0.01} \\

& ELF-PD          & \stk{120.58}{1.91} & \stk{5.49}{0.00} & \stk{88.29}{1.32} & \stk{5.45}{0.00} & \stk{63.73}{0.45} & \stk{5.39}{0.00} & \stk{59.82}{0.60} & \stk{5.36}{0.00} & \stk{58.45}{0.50} & \stk{5.35}{0.00} \\

& ELF-GAN &  \stk{86.06}{1.48} & \stk{5.31}{0.00} & \stk{52.88}{0.92} & \stk{5.17}{0.00} & \stk{41.08}{0.47} & \stk{5.07}{0.00} & \stk{36.06}{0.35} & \stk{5.02}{0.00} & \stk{32.86}{0.17} & \stk{5.00}{0.00} \\

& \textbf{ELF-MMD} & \stk{192.28}{1.08} & \stk{5.53}{0.00} & \stk{110.07}{0.69} & \stk{5.49}{0.00} & \stk{56.63}{0.13} & \stk{5.45}{0.00} & \stk{43.01}{0.20} & \stk{5.41}{0.00} & \stk{39.79}{0.29} & \stk{5.40}{0.00} \\

& \quad\quad \textit{+ IRD} & \stk{141.03}{0.73} & \stk{5.47}{0.00} & \stk{\textbf{78.11}}{0.54} & \stk{5.44}{0.00} & \stk{\textbf{47.33}}{0.29} & \stk{5.39}{0.00} & \stk{\textbf{38.83}}{0.23} & \stk{5.35}{0.00} & \stk{\textbf{35.75}}{0.15} & \stk{5.33}{0.00} \\

\cmidrule(lr){2-12}

& ELF$^{\star}$   & \stk{40.66}{8.85} & \stk{5.01}{0.21} & \stk{89.20}{9.55} & \stk{5.45}{0.03} & \stk{\textbf{79.37}}{2.61} & \stk{5.44}{0.03} & \stk{60.17}{4.76} & \stk{5.32}{0.04} & \stk{44.31}{1.91} & \stk{5.25}{0.01} \\

& FMLM$^{\star}$ &  \stk{170.26}{8.26} & \stk{5.49}{0.01} & \stk{136.92}{7.04} & \stk{5.51}{0.01} & \stk{116.84}{2.92} & \stk{5.48}{0.00} & \stk{90.66}{4.99} & \stk{5.37}{0.01} & \stk{100.94}{1.79} & \stk{5.34}{0.01} \\

& \textbf{ELF$^{\star}$-MMD} &  \stk{128.31}{1.29} & \stk{5.31}{0.00} & \stk{97.99}{0.79} & \stk{5.42}{0.00} & \stk{84.39}{0.35} & \stk{5.43}{0.00} & \stk{79.97}{0.29} & \stk{5.41}{0.00} & \stk{79.72}{0.85} & \stk{5.40}{0.00} \\

\midrule

\multirow{11}{*}[-2.4em]{GPT-2} & ELF             & \stk{17.86}{0.72} & \stk{2.34}{0.03} & \stk{164.19}{13.09} & \stk{4.20}{0.07} & \stk{320.55}{6.41} & \stk{5.16}{0.03} & \stk{132.07}{5.08} & \stk{5.29}{0.08} & \stk{57.54}{1.66} & \stk{5.33}{0.04} \\

& ELF-PD          & \stk{108.17}{3.07} & \stk{5.34}{0.00} & \stk{73.60}{1.86} & \stk{5.37}{0.00} & \stk{59.14}{1.14} & \stk{5.35}{0.00} & \stk{52.91}{0.75} & \stk{5.34}{0.00} & \stk{49.65}{0.60} & \stk{5.33}{0.00} \\

& ELF-GAN  & \stk{\textbf{103.15}}{1.54} & \stk{5.47}{0.00} & \stk{79.93}{0.94} & \stk{5.44}{0.00} & \stk{68.20}{0.54} & \stk{5.41}{0.00} & \stk{61.21}{0.67} & \stk{5.40}{0.00} & \stk{56.15}{0.20} & \stk{5.37}{0.00} \\

& \textbf{ELF-MMD} & \stk{127.77}{0.68} & \stk{5.39}{0.00} & \stk{85.61}{0.27} & \stk{5.45}{0.00} & \stk{56.52}{0.26} & \stk{5.45}{0.00} & \stk{44.34}{0.25} & \stk{5.44}{0.00} & \stk{40.01}{0.18} & \stk{5.43}{0.00} \\

& \quad\quad \textit{+ IRD} & \stk{114.32}{0.46} & \stk{5.36}{0.00} & \stk{\textbf{77.55}}{0.28} & \stk{5.42}{0.00} & \stk{\textbf{53.48}}{0.17} & \stk{5.43}{0.00} & \stk{\textbf{43.20}}{0.06} & \stk{5.42}{0.00} & \stk{\textbf{39.34}}{0.18} & \stk{5.42}{0.00} \\

\cmidrule(lr){2-12}

& ELF$^{\star}$    & \stk{84.45}{7.68} & \stk{4.13}{0.09} & \stk{156.48}{12.44} & \stk{5.40}{0.03} & \stk{88.43}{1.60} & \stk{5.42}{0.01} & \stk{66.65}{1.27} & \stk{5.41}{0.01} & \stk{66.78}{12.44} & \stk{5.40}{0.03}\\

& FMLM$^{\star}$  & \stk{95.25}{0.80} & \stk{5.45}{0.00} & \stk{75.82}{0.48} & \stk{5.41}{0.00} & \stk{57.32}{0.36} & \stk{5.37}{0.00} & \stk{59.93}{0.20} & \stk{5.40}{0.00} & \stk{55.17}{0.38} & \stk{5.38}{0.00} \\

& \textbf{ELF$^{\star}$-MMD}  & \stk{80.20}{1.15} & \stk{5.38}{0.00} & \stk{49.27}{0.93} & \stk{5.36}{0.00} & \stk{39.64}{0.77} & \stk{5.34}{0.00} & \stk{35.24}{0.48} & \stk{5.33}{0.00} & \stk{32.88}{0.21} & \stk{5.33}{0.00} \\

\bottomrule
\end{tabular}
\end{table}
\begin{table}[t!]
\centering
\caption{\textbf{Continuous DLMs on GSM8K.} For each number of sampling steps we report final-answer accuracy ($\%$). The results are averaged by 5 seeds.}
\label{tab:gsm_results}
\setlength{\tabcolsep}{4pt}
\small
\begin{tabular}{l |ccccccc}
\toprule
Method & 1 & 2 & 4 & 8 & 16 & 32 & 64 \\
\midrule
ELF & \ms{0.38}{0.05} & \ms{2.00}{0.33} & \ms{5.99}{0.20} & \ms{13.84}{0.84} & \ms{23.26}{0.47} & \ms{28.84}{0.91} & \ms{31.63}{0.94} \\
ELF-PD & \ms{\textbf{2.24}}{0.33} & \ms{6.31}{0.25} & \ms{15.95}{0.36} & \ms{23.52}{0.56} & \ms{27.16}{1.24} & \ms{28.32}{0.98} & \ms{28.48}{0.55} \\

ELF-GAN & \ms{0.73}{0.08} & \ms{3.38}{0.17} & \ms{9.73}{0.38} & \ms{20.27}{0.85} & \ms{29.11}{0.35} & \ms{32.77}{0.18} & \ms{34.36}{0.80} \\

\textbf{ELF-MMD} & \ms{0.71}{0.20} & \ms{4.00}{0.15} & \ms{14.21}{0.45} & \ms{27.49}{0.53} & \ms{34.16}{0.44} & \ms{34.28}{0.40} & \ms{35.24}{0.87} \\
\quad\quad\textit{+ IRD} & \ms{1.33}{0.12} & \ms{\textbf{7.46}}{0.67} & \ms{\textbf{20.76}}{0.77} & \ms{\textbf{32.46}}{0.68} & \ms{\textbf{35.62}}{0.89} & \ms{\textbf{35.97}}{0.62} & \ms{\textbf{36.30}}{0.60} \\

\cmidrule(lr){1-8}

ELF$^{\star}$ & \ms{0.60}{0.10} & \ms{4.14}{0.43} & \ms{12.69}{0.29} & \ms{25.31}{0.43} & \ms{30.25}{0.59} & \ms{32.40}{0.55} & \ms{33.59}{0.59} \\

\textbf{ELF$^{\star}$-MMD} & \ms{\textbf{0.79}}{0.12} & \ms{\textbf{4.44}}{0.28} & \ms{\textbf{15.33}}{0.75} & \ms{\textbf{28.52}}{0.41} & \ms{\textbf{33.50}}{0.65} & \ms{\textbf{34.40}}{0.51} & \ms{\textbf{35.07}}{0.37} \\

\cmidrule(lr){1-8}

FMLM${+}$     & \ms{0.26}{0.11} & \ms{0.71}{0.19} & \ms{5.88}{0.69} & \ms{16.04}{0.83} & \ms{25.63}{1.16} & \ms{30.86}{1.22} & \ms{32.37}{1.09} \\

\textbf{FMLM${+}$-MMD}     & \ms{\textbf{0.38}}{0.15} & \ms{\textbf{0.91}}{0.12} & \ms{\textbf{9.10}}{0.42} & \ms{\textbf{19.48}}{0.68} & \ms{\textbf{27.75}}{0.98} & \ms{\textbf{32.83}}{1.37} & \ms{\textbf{33.36}}{1.16} \\

\bottomrule
\end{tabular}
\end{table}

\clearpage

\clearpage
\begingroup
\captionsetup{hypcap=false}
\begin{table*}[p]
\centering
\captionsetup{skip=4pt}
\begin{minipage}{\linewidth}
\centering
\caption{\textbf{MDLM-MMD hyperparameters.} Settings for OWT and TinyGSM.}
\small
\renewcommand{\arraystretch}{1.0}
\setlength{\tabcolsep}{8pt}
\begin{tabular}{@{}lcc@{}}
    \toprule
    Hyperparameter
    & Unconditional (OWT)
    & Conditional (TinyGSM) \\
    \midrule

    \multicolumn{3}{@{}l}{\textbf{MMD training}} \\
    \addlinespace[2pt]
    Kernel
    & RBF & RBF \\
    Policy-gradient group size ($G$)
    & $4$ & $4$ \\
    MMD batch size ($B$)
    & $2$ & $2$ \\
    RBF bandwidth ($\sigma$)
    & $160$ & $90$ \\
    Feature layer
    & $4$ & $7$ \\
    \midrule

    \multicolumn{3}{@{}l}{\textbf{Optimization}} \\
    \addlinespace[2pt]
    Global batch size
    & $512$ & $512$ \\
    Optimizer
    & AdamW & AdamW \\
    Learning rate
    & $1 \times 10^{-4}$ & $1 \times 10^{-5}$ \\
    Training steps
    & $3{,}250$ & $7{,}000$ \\
    \bottomrule
\end{tabular}
\label{tab:mdlm_hyperparameters}
\end{minipage}
\par\vspace{18pt}
\begin{minipage}{\linewidth}
    \centering
    \caption{\textbf{ELF-MMD hyperparameters.} Settings for OWT and TinyGSM.}
    \small
    \renewcommand{\arraystretch}{1.0}
    \setlength{\tabcolsep}{8pt}

    \begin{tabular}{@{}lccc@{}}
        \toprule
        & \multicolumn{2}{c}{Unconditional (OWT)}
        & Conditional (TinyGSM) \\
        \cmidrule(l{\tabcolsep}r{\tabcolsep}){2-3}
        \cmidrule(l{\tabcolsep}){4-4}
        Hyperparameter
        & T5
        & GPT-2
        & GPT-2 \\
        \midrule

        \multicolumn{4}{@{}l}{\textbf{MMD training}} \\
        \addlinespace[2pt]
        Kernel
        & RBF & RBF & RBF \\
        MMD batch size ($B$)
        & $2$ & $4$ & $64$ \\
        RBF bandwidth ($\sigma$)
        & $50$ & $180$ & $3330$ \\
        Feature layer & 6 & 6 & 6 \\
        \midrule        

        \multicolumn{4}{@{}l}{\textbf{Conditioning and guidance}} \\
        \addlinespace[2pt]
        Bootstrapping $n$
        & $3$ & $4$ & $4$ \\
        Self-conditioning CFG range
        & $[0.5, 5]$ & $[0.5, 5]$ & $[1.0, 1.0]$ \\
        Noise resampling (\texttt{resample\_z})
        & \texttt{false} & \texttt{false} & \texttt{true} \\

        \midrule

        \multicolumn{4}{@{}l}{\textbf{Optimization}} \\
        \addlinespace[2pt]
        Global batch size
        & $64$ & $64$ & $2$ \\
        Optimizer
        & Muon & Muon & Muon \\
        Learning rate
        & $5 \times 10^{-5}$ & $5 \times 10^{-5}$ & $5 \times 10^{-5}$ \\
        Training steps
        & $10{,}000$ & $9{,}000$ & $7{,}000$ \\

        \midrule

        \multicolumn{4}{@{}l}{\textbf{IRD}} \\
        \addlinespace[2pt]
        Teacher rollout length $K$
        & $8$ & $8$ & $32$ \\
        Training steps
        & $5{,}500$ & $6{,}000$ & $20{,}000$ \\

        \bottomrule
        \label{tab:elf_hyperparameters}
    \end{tabular}
\end{minipage}
\par\vspace{18pt}
\begin{minipage}{\linewidth}
    \centering
    \caption{\textbf{DMax-MMD hyperparameters.} Settings for math and code post-training.}
    \small
    \renewcommand{\arraystretch}{1.0}
    \setlength{\tabcolsep}{8pt}

    \begin{tabular}{@{}lcc@{}}
        \toprule
        Hyperparameter
        & DMax-Math-MMD
        & DMax-Coder-MMD \\
        \midrule

        \multicolumn{3}{@{}l}{\textbf{MMD training}} \\
        \addlinespace[2pt]
        Kernel
        & RBF & RBF \\
        Policy-gradient group size ($G$)
        & $8$ & $8$ \\
        MMD batch size ($B$)
        & $2$ & $4$ \\
        RBF bandwidth ($\sigma$)
        & $10$ & $16$ \\
        Feature layer
        & $10$ & $10$ \\
        \midrule

        \multicolumn{3}{@{}l}{\textbf{Optimization}} \\
        \addlinespace[2pt]
        Optimizer
        & AdamW & AdamW \\
        Learning rate
        & $2 \times 10^{-6}$ & $2 \times 10^{-6}$ \\
        Adam coefficients ($\beta_1$, $\beta_2$)
        & $(0.9, 0.999)$ & $(0.9, 0.999)$ \\
        Training steps
        & $400$ & $400$ \\
        \bottomrule
    \end{tabular}
        \label{tab:dmax_hyperparameters}
\end{minipage}

\end{table*}
\endgroup

\clearpage

\begin{figure*}[!t]
\centering
\newsavebox{\trainingalgbox}
\begin{lrbox}{\trainingalgbox}
\begin{minipage}[t]{0.485\textwidth}
\lstset{lineskip=0.25pt}
\input{algos/training}
\end{minipage}
\end{lrbox}
\usebox{\trainingalgbox}
\hfill
\begin{minipage}[t][\dimexpr\ht\trainingalgbox+\dp\trainingalgbox\relax][s]{0.485\textwidth}
\input{algos/sampling}
\par\vspace{0.86mm}\vfill
\input{algos/ird}
\end{minipage}
\end{figure*}

\clearpage

\section{MDLM-MMD qualitative examples}
\label{app:examples_mdlm}

\subsection{Unconditional generation on OpenWebText}

We provide unconditional samples generated by MDLM-MMD on OWT at each sampling step, together with their generative perplexity and entropy. 

\definecolor{sampleheader}{RGB}{128,128,128}
\definecolor{samplebackground}{RGB}{245,245,245}

\newtcolorbox{sampleframe}[3]{
  enhanced,
  breakable,
  colback=samplebackground,
  colframe=sampleheader,
  colbacktitle=sampleheader,
  coltitle=white,
  boxrule=0.5pt,
  arc=1.5mm,
  boxsep=0pt,
  left=5pt, right=5pt, top=3pt, bottom=5pt,
  toptitle=3pt, bottomtitle=3pt,
  titlerule=0pt,
  fontupper=\rmfamily\small,
  fonttitle=\rmfamily\small\bfseries,
  title={Sampling Steps: #1\hfill
    {\mdseries gPPL: #2, Entropy: #3}},
  before upper={
    \setlength{\parindent}{0pt}
    \setlength{\parskip}{0pt}
  },
  before skip=8pt,
  after skip=8pt
}
\begin{sampleframe}{2}{708.40}{5.42}
\ldots{} who because of her alleged offenses, systematically within systematically systematically abused unlawfully, inflicting physically different physical and psychological abuses against her,  even activistsited ally herself in," comfortable of her misconduct among many in which she believed she as hoped enough could bring leverage to her malicious, ", she explained while exploring heat her "far far more," certain allies and said of her<|endoftext|>andails in placing her within a systematically pervasive government bureaucracy could harm her seeking system premiums game caused, would, she whether, misconduct demands, or violate her ” Medicaid reimburse herselfor enabled and and carefully evidence evidence if she had violated unlawfully transgress illegally or she could be guaranteeing being email precisely this, there, request to themselves \ldots{}
\end{sampleframe}

\begin{sampleframe}{4}{120.65}{5.43}
\ldots{} her transfer, and that, despite her assurances from her efforts, she endured different physical and psychological trauma than her,  even activists who had been in the consequences of government surveillance. Although she was repeatedly harassed and subjected to abusive abuse throughout her experiences, she repeatedly explained to investigators that despite had "reasonable suspicion," certain authorities lawyers said that she had not committed any legal wrongdoing within any law or government authoritythat she was seeking information about. "I would ask she whether any government demands for removal remove her” would “be enabled and and carefully evidence-free investigation,” as “she could have herself being released precisely … were there was request to remove her" But was her request be handled properly?<|endoftext|>Jennifer Lawyer, \ldots{}
\end{sampleframe}

\begin{sampleframe}{8}{45.04}{5.42}
\ldots{} politically motivated information fraud," she said, "we would be in a very different position after working with the government." But activists concern that the government's attempts to use this power could make it hard to find Kim Kim for alleged crimes. "If they're weak, they'll be able to arrest her in certain cases," said Lila Wong, an environmental activist, who is asking the government to stop foreigners from seeking information about her. But Kim asked she whether she would be able to pursue her case if her accusers were enabled and presented with evidence without a subpoena Wong suggested that Kim could easily come up with precisely which countries there would request for her information. InIn response she stressed that she had not been advising Kim in her foreign ministry for more than three years, and that she had acted \ldots{}
\end{sampleframe}

\begin{sampleframe}{16}{28.58}{5.43}
\ldots{} lla Darev has demanded compensation and compensation worth £175m over allegations allegations against her parents at a Moscow university. But Ms Dare said she had sympathy with the plight of every girl who receives education training at Moscow university, and would urge them to take the necessary steps necessary to protect them from from future campus accidents. She said: "Moscow university is fully aware of the allegations against me and we will play a critical role in ensuring that students are fully protected. "We are confident of the compliance reforms that we have enabled and have no evidence of further abuse or exploitation. "We are working to ensure that we believe that there is no inappropriate conduct taking place." She said she had not been used for sexual abuse by other staff at students over the last three years and had \ldots{}
\end{sampleframe}

\begin{sampleframe}{32}{21.52}{5.44}
\ldots{} At the very least, over the last six years, we have been pleased that Secretary Clinton has worked tirelessly to build an education and training infrastructure that will enable our Nation to meet its goal of providing high quality education for all children.”\par\vspace{2bp}\noindent Clinton’s campaign has not so far commented publicly about her use of the server. Last week, campaign spokesman Henry Johnson issued a few statements about the candidate’s use of the private email server:\par\vspace{2bp}\noindent “The Department does not allow the business interests of any individual person to play any role in the development, preparation, production, distribution, sale, purchase of, or other selling of controlled substances directly or indirectly by any employees or employees of the Department or its employees.” \ldots{}
\end{sampleframe}

\subsection{Conditional generation on TinyGSM}
We provide examples for MDLM-MMD on TinyGSM.
\begin{gsmcode}[unbreakable,fonttitle=\rmfamily\small\bfseries,before skip=5pt,after skip=0pt,left=4pt,right=4pt,top=2pt,bottom=2pt,toptitle=1pt,bottomtitle=1pt]{2}{0}{Error}
def simple_math_problem() -> int:
    '''
    There are three trees in Eddy's backyard.
    The shortest tree has a height of 6 feet, and the second tree has a height of 5 feet more than the shortest tree.
    The height of the tallest tree is twice the height of the two trees combined.
    How tall is the tallest tree?
    '''
    height_tree_1 = 6
    height_tree_2 = height_tree_1 + 5
    height_tree_3 = 2 * ((*@\gsmerror{\_\_tree\_}@*) + (*@\gsmerror{height\_\_\_2}@*))
    (*@\gsmerror{=\space{}=\_tree\_3}@*)
    return result
\end{gsmcode}
\begin{gsmcode}[unbreakable,fonttitle=\rmfamily\small\bfseries,before skip=5pt,after skip=0pt,left=4pt,right=4pt,top=2pt,bottom=2pt,toptitle=1pt,bottomtitle=1pt]{3}{0}{Error}
def simple_math_problem() -> int:
    '''
    There are three trees in Eddy's backyard.
    The shortest tree has a height of 6 feet, and the second tree has a height of 5 feet more than the shortest tree.
    The height of the tallest tree is twice the height of the two trees combined.
    How tall is the tallest tree?
    '''
    shortest_tree_height = 6
    second_tree_height = shortest_tree_height + 5
    tallest_tree_height = 2 * ((*@\gsmerror{short\_tree\_height}@*) + (*@\gsmcorrect{second\_tree\_height}@*))
    (*@\gsmcorrect{result\space{}=}@*) (*@\gsmerror{tallesttreetree\_height}@*)
    return result
\end{gsmcode}
\begin{gsmcode}[unbreakable,fonttitle=\rmfamily\small\bfseries,before skip=5pt,after skip=0pt,left=4pt,right=4pt,top=2pt,bottom=2pt,toptitle=1pt,bottomtitle=1pt]{4}{1}{34}
def simple_math_problem() -> int:
    '''
    There are three trees in Eddy's backyard.
    The shortest tree has a height of 6 feet, and the second tree has a height of 5 feet more than the shortest tree.
    The height of the tallest tree is twice the height of the two trees combined.
    How tall is the tallest tree?
    '''
    shortest_tree_height = 6
    second_tree_height = shortest_tree_height + 5
    tallest_tree_height = ((*@\gsmcorrect{shortest\_tree\_height}@*) + second_tree_height)* 2
    result = (*@\gsmcorrect{tallest\_tree\_height}@*)
    return result
\end{gsmcode}

\clearpage
\section{ELF-MMD qualitative examples}
\label{app:examples_elf}

\subsection{Unconditional generation on OpenWebText}

We provide unconditional samples generated by ELF-MMD on OWT at each sampling step, together with their generative perplexity and entropy. 

\subsubsection{T5 Encoder}

\begin{sampleframe}{2}{128.72}{5.49}
\ldots{} a regular factor in the procedures are surveys procedures, but the absence of a document which on the matter is the Commission, Prime Minister Dangg Ghai made it clear that the president and Nuember of Guinea have refused to change the matter and can not to do so. "It's a serious problem to my country's commitment to provide assistance to all people who are not able to handle anything involving gender, gender, racism and rape," he said. "We can conduct crimeness without fear of flight." When the DaoF-I published on the resolution earlier this month, human music organization TIEFES said the change paints are not intended to help people  people because the country wants to show themselves to recognize their they are their \ldots{}
\end{sampleframe}

\begin{sampleframe}{4}{70.51}{5.50}
\ldots{} program, but with lack of a official release on the matter by theFA, Prime Minister Jenar Ghai made it clear that the president and the President of Uganda have refused to address the matter and agreeing to do so. "It's a serious breach of my country's commitment to provide justice to all who have not been unable to address issues such as race, religion, alcohol and rapicide," he said. "We can continue doing so without fear of fire." While the Nationalation of Marine Sciences reported on the matter earlier this month, human rights group UIIEES said the change of standards is not targeted to the handicaps, because the group wants to show their identity with their speakers and their languages. It also seeks to distribute safety information to \ldots{}
\end{sampleframe}

\begin{sampleframe}{8}{43.71}{5.43}
\ldots{} but with lack of a formal statement on the matter, Nudeter Prime Minister Inar Ghasi made it clear that the president and the President of Afghanistan have refused to address the issue and pledged to do so. "It's a fundamental part of my country's mission to give justice to anyone who may not be able to accept issues such as race, income, alcohol and facy," he said. "We can continue doing so without fear of fire." While the National Academy for Marine Science reported on the matter earlier this month, human rights group UIAIUK says the change of rules is not limited to the homosexuals, as the group wants to demonstrate their confidence in their languages and their languages. It also seeks to distribute this information to people in their area who are severely homeless or deprived of medical care and those living in the Africa Park Golf Club, which \ldots{}
\end{sampleframe}

\begin{sampleframe}{16}{38.22}{5.41}
\ldots{} new backgrounds program, but with lack of a formal statement on the matter, Nudeter press secretary Inar Ghati made it clear that the president and the President of Afghanistan have refused to address the issue and pledged to do so. “It’s a fundamental part of my country’s history to give justice to those who may not be able to accept things such as race, alcohol, alcohol and piracy,” he said. “We can continue doing so without fear of violence.” While the National Academy for Marine Science reported on the matter earlier this month, human rights group UIERAUK says the backgrounds program is not limited to Africanns, as the group wants to demonstrate their faith in their religion and their cultures. It also seeks to relay this information to people in the area who are severely disabled or deprived of medical care and those participating in the AFC, which \ldots{}
\end{sampleframe}

\begin{sampleframe}{32}{35.69}{5.40}
\ldots{} the backgrounds program, but with lack of a formal statement on the matter, Nudeter press secretary Inar Ghati made it clear that the president and the President of Afghanistan have refused to address the issue and pledged to do so. “It’s a fundamental part of our country’s history to give justice to those who may not be able to accept things such as race, drug, alcohol and piracy,” he said. “We can continue doing so without fear of violence.” While the National Academy for Marine Science reported on the matter earlier this month, human rights group UIERAUK said the backgrounds program is not available to Alaskans, because the group wants to demonstrate their faith in their religion and their cultures. It also seeks to extend its coverage to people in the area who are severely disabled or deprived of medical care and those participating in the AlaskaF Football Game, which can be played on a daily basis at 9 a.m. At the time, Gagibo said the backgrounds program \ldots{}
\end{sampleframe}

\subsubsection{GPT-2 Encoder}

\begin{sampleframe}{2}{91.65}{5.37}
\ldots{} The Bitcoin of Pledge will truly support our money together. The capitalists and Bitcoiners distribute this precious currency to legitimate middle-class customers, and as some of us, we can find, try and move it into our supply.carts, rent, shop, rent them, and the same products are not worthless. [FORDREP ] THE Purpose of Crypto Party

The Party is not invention and belongs to totalitarian rulers who controlled their society. From existing with medieval wars, the Noms, and the outbreak of the Naziism, all of them are dead. When the fascists had taken over this realm of brutalism and in the ensuing a period of warfare, the Fascist Party was again resurrected – the Green Party, the Republicanists of Fascists, the NX Party, and the Green Society. 4. Two Multi Kinders ( of Silver, sized for \$2250 (US\$120 USD) and a NCoin-Gold-Coin Gold Shield Case for \$1350. 4. FRatch of the Piece of “( |YXER” PIX. To provide the more composition and \ldots{}
\end{sampleframe}

\begin{sampleframe}{4}{58.36}{5.43}
\ldots{} 3. The Price of Gold will dramatically improve our financial system. Bankaires and venders deliver this valuable commodity to our middle-class customers, and as many of us, we can buy, store and transfer it to our home. Fund stocks, buy, sell, and silver together and ensure great values are no hardship. Reward=JDY!

The Purpose of STEERINGER is nothing true, than the totalitarianists that controlled our society. From the first three years of the Namine, to the end of the Great War, all of them are dead. When the communists had taken over this world by ruthless force, and turned it four years of warfare, the NFA Group became several identities – the Green Party, the Nance of Thoms, the NAR Party, and the Green Party. 4. Two Rock Party Cards come in size, starting at \$2595 (US \$240 USD) and a NEM-In-Case / Shield Case for \$2895. 4) A sleeve of the item called “SLEERYER” Sleeve. To provide further more functionality \ldots{}
\end{sampleframe}

\begin{sampleframe}{8}{41.71}{5.40}
\ldots{} Private investors and large investors add this invaluable metal to our middle-class purchases, and as many of us, we should purchase, store and incorporate it into our home. Use it, buy, sell, refribute, and ensure your purchases go without hassle, quality in your time. Feder is nothing more more than the totalitarian forces that control our society.

From the first six years of the NED, to the beginning of the Great War, all of them are dead. When the Nazis had taken over this world by ruthless force and after more than two decades of war, the NY Coalition came in groups: the Buyer, the Nition of Pools, the NEM Group, and the Pioneer. 1) Two Solder Packs come in series, starting at \$20 USD (about \$70 USD) and a NEM-PlusxSize/Cube Pack for \$35 USD. 2) Each box will come with a “SOLDER ROCKER” Sleeve. To provide even \ldots{}
\end{sampleframe}

\begin{sampleframe}{16}{37.06}{5.37}
\ldots{} 3) The uptake of silver will greatly improve our food economy. Big businesses and main grains make a significant part in our mid-day diets, and as all of us, we should consume, store and incorporate silver in our food. Buy it, sell, sell, redistribute, and redistribute it all without any extra charge in our diet.

Founder is nothing much more than the Satanic forces that control our society. From the first six years of the Noney War to the end of the World War, all of them are over. After the Russians had taken over the world in 1844, after more than a decade of war, the Nurity Coalition came to form: the Solder, the Nition of Pounds, the Nurity Group, and the Solder. 4) Two Solder Packs available in package, starting at \$20 Silver ( for \$60 USD) and a NEM Sleeve (the Solder Bundle for \$60 USD). 4) Each box will come with a “SOLDER WATCHER” SIZE. To allow \ldots{}
\end{sampleframe}

\begin{sampleframe}{32}{36.97}{5.34}
\ldots{} Silver is not a monohydritic metal. It is made of elements like copper and copper. It is an essential commodity and will have an impact in health for those who work in the energy industry. 3) The availability of silver will significantly improve our food economy. Fibers and silver are playing a vital role in our mid-day diet, and as many of us, we regularly consume, store and incorporate them into our food. Solders, blend, bottle, resele, and unitizers come without an extra ingredient in our food. 

Cryer is nothing much more than the Satanic forces that controlled our society. From the first six years of the Noney War to the end of the Great War, all of them are dead. After the Russians had taken over the world in 1844, after more than a century of warfare, the Nurity Coalition came to form: the Solder, the Nition of Pints, the Nurity Society, and the Solder. 4) Two Solder Packs available in package, starting at \$60 (total of \$69 USD) and a NEM Sleeve (the Solder Bundle costs \$48 USD). 4) Each unit will come with a “SOLDER CLICKER” Sleeve. To allow easier product creation and \ldots{}
\end{sampleframe}

\subsection{Conditional generation on TinyGSM}

We provide examples for ELF-MMD on TinyGSM.

\begin{gsmcode}[unbreakable,fonttitle=\rmfamily\small\bfseries,before skip=5pt,after skip=0pt,left=4pt,right=4pt,top=2pt,bottom=2pt,toptitle=1pt,bottomtitle=1pt]{1}{0}{Error}
def simple_math_problem() -> int:
    '''
    Lloyd earns $10 an hour on Math tutoring.
    And he tutored 5 hours for the first week and 8 hours for the second week.
    How much did he earn for the first two weeks?
    '''
    hourly_rate = 10
    hours_first_week = 5
    hours_second_week = 8
   (*@\gsmerror{total\space{}=40\space{}and\space{}hours\_first\_week\space{}*\space{}hours\_week}@*)
      (*@\gsmerror{total4033233}@*)
(*@\gsmerror{3}@*)
(*@\gsmerror{40}@*)
   
 result (*@\gsmerror{and\space{}\space{}and}@*) result

\end{gsmcode}

\begin{gsmcode}[unbreakable,fonttitle=\rmfamily\small\bfseries,before skip=5pt,after skip=0pt,left=4pt,right=4pt,top=2pt,bottom=2pt,toptitle=1pt,bottomtitle=1pt]{2}{0}{Error}
def simple_math_problem() -> int:
    '''
    Lloyd earns $10 an hour on Math tutoring.
    I He tutored 5 hours for the first week and 8 hours for the second week.
    How much did he earn for the first two weeks?
    '''
    hourly_rate = 10
    hours_first_week = 5
    hours_second_week = 8
    (*@\gsmcorrect{total\_hours\space{}=\space{}hours\_first\_week\space{}+\space{}hours\_second\_week}@*)
    (*@\gsmcorrect{total\_earnings}\space{}=\space{}\gsmcorrect{hourly\_rate}@*) * (*@\gsmerror{hourly\_rate}@*)
    (*@\gsmcorrect{result\space{}=\space{}total\_earnings}@*)
    (*@\gsmcorrect{return}@*) (*@\gsmerror{resultA}@*)
\end{gsmcode}

\begin{gsmcode}[unbreakable,fonttitle=\rmfamily\small\bfseries,before skip=5pt,after skip=0pt,left=4pt,right=4pt,top=2pt,bottom=2pt,toptitle=1pt,bottomtitle=1pt]{4}{1}{130}
def simple_math_problem() -> int:
    '''
    Lloyd earns $10 an hour on Math tutoring.
     He tutored 5 hours for the first week and 8 hours for the second week.
    How much did he earn for the first two weeks?
    '''
    hourly_rate = 10
    hours_first_week = 5
    hours_second_week = 8
    total_hours = hours_first_week + hours_second_week
    total_earnings = hourly_rate * (*@\gsmcorrect{total\_hours}@*)
    result = total_earnings
    return (*@\gsmcorrect{result}@*)
\end{gsmcode}

\end{document}